\documentclass{article}
\usepackage[final,main]{neurips_2026}
\usepackage[utf8]{inputenc}
\usepackage[T1]{fontenc}
\usepackage{courier}
\usepackage{amsmath,amssymb,amsthm,bm}
\usepackage{booktabs,multirow,xcolor,colortbl}
\usepackage{float}
\usepackage{graphicx}
\floatstyle{ruled}
\newfloat{algorithm}{tbp}{loa}
\floatname{algorithm}{Algorithm}
\usepackage{pgfplots}
\usetikzlibrary{positioning,shapes.geometric,arrows.meta,calc,fit,backgrounds,patterns}
\pgfplotsset{compat=1.18}
\usepackage{subcaption}
\usepackage[hidelinks]{hyperref}
\usepackage{cleveref}
\usepackage{enumitem}
\usepackage{placeins}
\usepackage{diagbox}
\usepackage{array}
\newcolumntype{P}[1]{>{\raggedright\arraybackslash}p{#1}}

\newtheorem{theorem}{Theorem}
\newtheorem{proposition}[theorem]{Proposition}
\newtheorem{lemma}[theorem]{Lemma}

\newtheorem{corollary}[theorem]{Corollary}

\newcommand{\E}{\mathbb{E}}
\newcommand{\R}{\mathbb{R}}
\newcommand{\cS}{\mathcal{S}}

\newcommand{\cL}{\mathcal{L}}

\newcommand{\Voc}{\mathrm{VoC}}
\newcommand{\dmax}{d_{\max}}
\newcommand{\dstar}{d^{*}}

\definecolor{aic}{HTML}{C0392B}
\definecolor{oursrow}{gray}{0.93}

\newcommand{\TaskControl}{HalfCheetah-v4}
\newcommand{\TaskManip}{ManiSkill\,PickCube-v1}
\newcommand{\TaskStack}{ManiSkill\,StackCube-v1}
\newcommand{\TaskNav}{AntMaze\,(main)}
\newcommand{\TaskNavNeg}{AntMaze\,(low-het.)}
\newcommand{\TaskLLM}{GPT-2 on GLUE-RTE}
\newcommand{\TaskRare}{RareEvent-StackCube-Misalign}
\newcommand{\TaskMGDist}{MiniGrid-DistShift2}
\newcommand{\TaskMGFour}{MiniGrid-FourRooms}
\newcommand{\TaskOGScene}{OGBench scene-play-v0}
\newcommand{\TaskOGCube}{OGBench cube-double-play-v0}
\newcommand{\TaskOGPuzzle}{OGBench puzzle-4x4-play-v0}

\makeatletter
\renewcommand{\paragraph}{\@startsection{paragraph}{4}{\z@}{0pt}{-1em}{\normalsize\bfseries}}
\makeatother

\title{\textbf{Learning When to Think:
Adaptive Internal Computation for Reinforcement Learning}}
\author{Rohit Kumar Salla \qquad Simon Stepputtis
\\ Virginia Polytechnic Institute and State University \\ \texttt{\{rohits25, stepputtis\}@vt.edu}}
\date{}

\begin{document}
\maketitle

\begin{abstract}
Iterative neural policies usually spend the same computation on every decision, although the benefit of additional processing varies across states.
We introduce Adaptive Internal Computation (AIC), an iterative policy that learns when to stop refining its latent representation and act. 
In AIC, we train action selection and stopping jointly, an expected-depth penalty values computation and intermediate value predictions provide auxiliary supervision. 
We show that when only some states benefit from further refinement, an ideal state-dependent allocator outperforms every state-independent stopping schedule at the same expected depth.
With at most eight refinement steps per decision, AIC executes 3.3 steps on average across twelve benchmark tasks and lowers per-decision latency by 31--44\% relative to a fixed eight-step policy on four control and robotics tasks.
Its mean reward exceeds the best fixed-depth policy on seven tasks and trails it on four.
Ablating adaptive stopping and value supervision separately shows that stopping accounts for most of the reward gain on PickCube and AntMaze, where it also executes more refinement steps and that supervision accounts for most of it on RareEvent.
Project page: https://rohitsalla.github.io/aic-neurips2026/
\end{abstract}

\section{Introduction}
\label{sec:intro}
Neural policies typically use a fixed computation graph to map each state to an action distribution.
Iterative policies instead repeatedly refine a representation of the current state before selecting an action, reusing one block so that computation grows without adding parameters~\citep{ghugare2026computation}.
These iterations process the current decision before the next environment transition; however, the number of effective iterations, or refinement depth, varies from one decision to the next.
Routine motion may need little processing, whereas recovery from an unexpected contact benefits from additional iterations.
When computation is costly, using the same depth everywhere spends too much on some states and too little on others, leading to the central question of how a policy can learn, at each state, whether further refinement is worth its cost.

Learned stopping provides a mechanism for making this allocation.
ACT and PonderNet learn how many iterations to use for an input through prediction objectives and computation regularization~\citep{graves2016adaptive,banino2021pondernet}; learned metacontrol uses RL to choose internal computations~\citep{hamrick2017metacontrol}.
For iterative policies, \citet{ghugare2026computation} establish benefits of recurrent computation at fixed depth and POGP~\citep{salla2026pogp} stops diffusion denoising once predicted value gains become small; learning the stopping decision itself from environment return remains open.
Here, stopping determines which action is executed and therefore affects subsequent states and rewards.
Learning when to stop requires assigning credit to that decision jointly with learning how to act.

To address this problem, we introduce Adaptive Internal Computation (AIC), which makes action selection and stopping two decisions of the same policy.
A halting head reads the current representation and chooses whether to refine it further or use it to act now.
Proximal policy optimization (PPO) updates the probabilities of both actions and stopping depths, while an expected-depth penalty charges for computation.
Intermediate value predictions receive a shared rollout target as auxiliary supervision (\Cref{sec:method}).
Our contribution has three parts. 
First, we formulate a joint distribution over stopping depth and action so that the environment return trains both decisions.
Second, we show that when only some states benefit from further refinement, an ideal state-dependent policy outperforms every state-independent schedule at the same expected depth.
Third, we evaluate AIC against fixed-depth and adaptive baselines, with a training-data control and an ablation of adaptive stopping and value supervision.

\section{Related Work}
\label{sec:related}
AIC builds on methods that adapt computation to the input, on prior uses of extra computation inside policies and on early stopping in diffusion policies; we also draw on methods for attributing performance gains.

\paragraph{Adaptive computation.}
Adaptive-computation methods let a network vary how much it computes for each input.
Adaptive Computation Time (ACT) adds a halting unit to a recurrent network and trains it through the task loss and a ponder cost~\citep{graves2016adaptive}, a mechanism that Universal Transformers apply per position~\citep{dehghani2018universal}.
PonderNet instead learns a conditional stopping probability at each step, trained through an expected prediction loss and a regularizer toward a geometric prior~\citep{banino2021pondernet}.
Early-exit networks attach classifiers to intermediate layers and stop once a prediction is confident~\citep{teerapittayanon2016branchynet,huang2017multi}, while routing methods learn to skip individual layers or blocks for a given input~\citep{wang2018skipnet,wu2018blockdrop,raposo2024mixture}.
AIC uses these established stopping primitives to select a depth-specific action policy.
Its RL formulation includes both action and stopping probabilities in the PPO ratio, together with an expected-depth penalty and intermediate value supervision (\Cref{app:precursor}).

\paragraph{Computation in policies.}
\citet{ghugare2026computation} studies how recurrent computation improves fixed-depth policies and identifies state-dependent depth within one policy as an open direction.
In contrast, AIC learns that depth decision directly, examining the allocation opportunity at equal expected computation.
While metareasoning and learned metacontrol already optimize computation choices~\citep{hay2014selecting,hamrick2017metacontrol}, AIC applies this idea to refinement inside the action policy, without requiring explicit transition-model queries.
Planning and search methods also spend additional computation on each decision, by expanding a search tree or unrolling a learned model~\citep{tamar2016value,farquhar2017treeqn,oh2017value,guez2018learning,silver2016mastering,schrittwieser2020mastering}; however, AIC instead refines a latent representation and queries no model.


\paragraph{Stopping in diffusion policies.}
POGP learns prefix values by bootstrapping along a diffusion denoising chain and stops when successive predicted gains are small~\citep{salla2026pogp}, so its value predictions supply the stopping rule without a halting network.
AIC instead trains a halting head through the joint policy ratio and a depth penalty; its per-depth critics are auxiliary and are not read at inference (\Cref{app:calibration}).
Stopping therefore does not rely on value increments or on the diminishing-returns condition under which a first-crossing rule is optimal (\Cref{app:halt-derivation}).

\paragraph{Attributing gains.}
Intermediate supervision can improve representations~\citep{lee2015deeply} and changing the policy can change the data it collects.
Our ablation and data comparisons examine these effects, while early-exit interventions test sensitivity to truncation~\citep{vig2020investigating,morcos2018importance} (\Cref{sec:causal}).

\section{Method: Adaptive Internal Computation}
\label{sec:method}

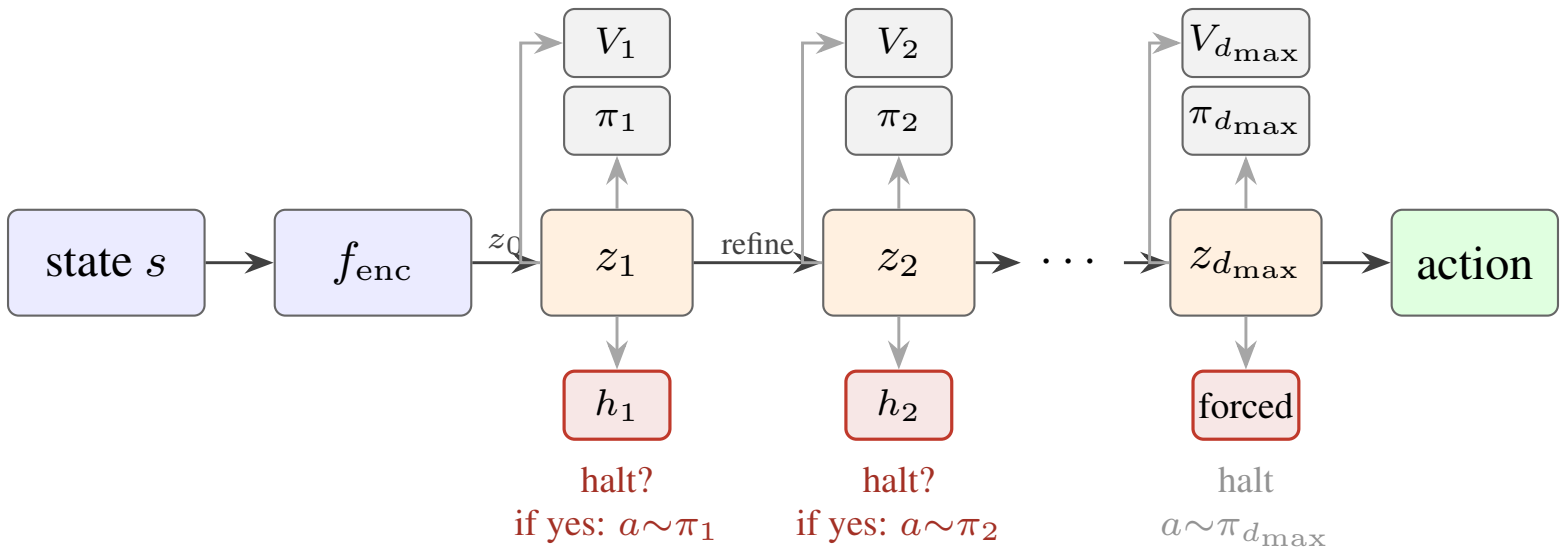
\begin{figure}[!ht]
\centering
\begin{tikzpicture}[
    >=Stealth, font=\small,
    enc/.style={draw=black!60, fill=blue!8, rounded corners=2pt, minimum width=1.3cm, minimum height=0.7cm, align=center},
    ref/.style={draw=black!60, fill=orange!12, rounded corners=2pt, minimum width=1.0cm, minimum height=0.7cm, align=center},
    pihead/.style={draw=black!60, fill=gray!10, rounded corners=2pt, minimum width=0.7cm, minimum height=0.45cm, font=\scriptsize, align=center, inner sep=1pt},
    vhead/.style={draw=black!60, fill=gray!10, rounded corners=2pt, minimum width=0.7cm, minimum height=0.45cm, font=\scriptsize, align=center, inner sep=1pt},
    halt/.style={draw=aic, fill=aic!12, rounded corners=2pt, minimum width=0.7cm, minimum height=0.45cm, font=\scriptsize, align=center, inner sep=1pt, line width=0.6pt},
    act/.style={draw=black!60, fill=green!12, rounded corners=2pt, minimum width=1.1cm, minimum height=0.7cm, align=center},
    arrow/.style={->, thick, black!75, line width=0.55pt},
    voc/.style={->, semithick, aic!85!black, dashed, line width=0.55pt},
]
\node[enc] (s) at (0,0) {state $s$};
\node[enc, right=0.45cm of s] (enc) {$f_{\mathrm{enc}}$};
\draw[arrow] (s) -- (enc);
\node[ref, right=0.45cm of enc] (z1) {$z_1$};
\node[ref, right=0.85cm of z1] (z2) {$z_2$};
\node[right=0.30cm of z2] (dots) {$\cdots$};
\node[ref, right=0.30cm of dots] (zd) {$z_{\dmax}$};
\node[act, right=0.45cm of zd] (action) {action};
\draw[arrow] (enc) -- node[above, font=\tiny, yshift=-2pt] {$z_0$} (z1);
\draw[arrow] (z1) -- node[above, font=\tiny, yshift=-2pt] {refine} (z2);
\draw[arrow] (z2) -- (dots);
\draw[arrow] (dots) -- (zd);
\draw[arrow] (zd) -- (action);
\node[pihead, above=0.35cm of z1] (pi1) {$\pi_1$};
\node[vhead, above=0.05cm of pi1] (v1) {$V_1$};
\draw[arrow, gray!70] (z1.north) -- (pi1.south);
\draw[arrow, gray!70] (z1.west) -- ++(-.13,0) |- (v1.west);
\node[pihead, above=0.35cm of z2] (pi2) {$\pi_2$};
\node[vhead, above=0.05cm of pi2] (v2) {$V_2$};
\draw[arrow, gray!70] (z2.north) -- (pi2.south);
\draw[arrow, gray!70] (z2.west) -- ++(-.13,0) |- (v2.west);
\node[pihead, above=0.35cm of zd] (pid) {$\pi_{\dmax}$};
\node[vhead, above=0.05cm of pid] (vd) {$V_{\dmax}$};
\draw[arrow, gray!70] (zd.north) -- (pid.south);
\draw[arrow, gray!70] (zd.west) -- ++(-.13,0) |- (vd.west);
\node[halt, below=0.35cm of z1] (h1) {$h_1$};
\node[halt, below=0.35cm of z2] (h2) {$h_2$};
\node[halt, below=0.35cm of zd] (hd) {forced};
\draw[arrow, gray!70] (z1.south) -- (h1.north);
\draw[arrow, gray!70] (z2.south) -- (h2.north);
\draw[arrow, gray!70] (zd.south) -- (hd.north);
\node[font=\scriptsize, color=aic!85!black, below=0.05cm of h1, align=center] {halt? \\ if yes: $a{\sim}\pi_1$};
\node[font=\scriptsize, color=aic!85!black, below=0.05cm of h2, align=center] {halt? \\ if yes: $a{\sim}\pi_2$};
\node[font=\scriptsize, gray!85, below=0.05cm of hd, align=center] {halt \\ $a{\sim}\pi_{\dmax}$};
\end{tikzpicture}
\caption{\textbf{AIC architecture.}
The encoder produces $z_0$; a shared residual MLP refines $z_d$.
Each depth has action and value heads, with a learned halting head before the forced cap.
The action and halting heads learn through the joint policy objective; value heads provide auxiliary shared-return supervision.}
\label{fig:architecture}
\end{figure}

We consider a policy that interacts with an environment in discrete steps: at step $t$ it observes a state $s_t$, executes an action $a_t$ and receives a reward $r_t$ and it aims to maximize expected discounted return.
An iterative policy does not map $s_t$ to an action in a single pass.
It refines a latent representation of $s_t$ over internal steps $d=1,\ldots,\dmax$, during which the environment supplies no new observation or reward.
Each internal step costs computation and the number of useful steps differs across states.
The problem is to learn, together with the actions, a stopping depth $\tau_t\in\{1,\ldots,\dmax\}$ for each state that keeps return high and expected depth low.
AIC treats $(\tau_t,a_t)$ as one decision of a joint policy, so the return that follows trains both choices (\Cref{fig:architecture}).

\paragraph{Architecture.}
An encoder $f_{\mathrm{enc}}: \cS\to\R^{d_z}$ maps the state to $z_0=f_{\mathrm{enc}}(s)$.
It uses a two-layer MLP for proprioception, a small CNN for grid states, or a frozen GPT-2~\citep{radford2019language} for language inputs.
A residual block then applies $z_d=z_{d-1}+\mathrm{MLP}(z_{d-1})$ for $d=1,\ldots,\dmax$, reusing its parameters at every depth.
Three lightweight heads read each $z_d$: an action distribution $\pi_d$, a value prediction $V_d$ and a conditional stopping probability $h_d\in[0,1]$ for $d<\dmax$.
Stopping at the cap is forced.
The main configuration uses $d_z=256$ and $\dmax=8$ (\Cref{fig:architecture,app:arch}).

Each $z_d$ represents the current observation after $d$ transformations.
The heads read this representation directly, with value supervision updating the features shared by action selection and halting during training.
The value outputs themselves are auxiliary: the halting head takes $z_d$ as its input.

\paragraph{Executing one decision.}
The policy encodes $s_t$, computes $z_1$ and either stops to draw $a_t\sim\pi_1$ or refines again.
This repeats until the selected depth or forced cap.
It then sends one action to the environment; internal refinement supplies no new observations or intermediate actions.

\paragraph{Training and evaluation.}
Training samples Bernoulli stopping decisions and evaluates all depths for auxiliary value supervision.
At evaluation, refinement ends at the first $d<\dmax$ with $h_d>0.5$, or at the cap and an action is drawn from $\pi_d$.
We measure evaluation return and depth under this threshold rule; \Cref{app:deployment-walkthrough} explains its relationship to stochastic training.

\subsection{Joint Policy and Training}
\label{sec:method-voc}

For the return to train stopping as well as acting, the stopping depth must enter the policy's probability model.
We therefore define a joint distribution over depth and action, built from the per-depth stopping probabilities.
Let $h_{\theta,d}(s)$ be the probability of stopping upon reaching depth $d$, so that the probability of reaching $d$ is $S_{\theta,d}(s)=\prod_{j<d}(1-h_{\theta,j}(s))$, with an empty product of one.
The stopping depth and action then follow

\begin{equation}
q_\theta(d\mid s)=
\begin{cases}S_{\theta,d}(s)h_{\theta,d}(s),&d<\dmax,\\
S_{\theta,\dmax}(s),&d=\dmax,
\end{cases}
\quad
p_\theta(a,d\mid s)=q_\theta(d\mid s)\pi_{\theta,d}(a\mid z_{\theta,d}(s)).
\label{eq:joint-policy}
\end{equation}
The cap receives the probability of surviving every earlier stopping decision. This makes $q_\theta$ a normalized distribution and gives $\E[\tau\mid s]=\sum_d S_{\theta,d}(s)$ (\Cref{lem:stopping-law}).

The distinction between $h_d$ and $q_\theta(d\mid s)$ matters: $h_d$ is conditional on having reached depth $d$, whereas $q_\theta$ includes every earlier decision to continue.
For example, stopping at depth two requires continuing at depth one and then stopping, so its probability is $(1-h_1)h_2$ when $2<\dmax$.
The joint log-probability consequently contains the continuation decisions, the final stopping decision when it is not forced and the selected action's log-probability.

For a sampled pair, a positive advantage means that the estimated return exceeds the rollout baseline.
The actor objective favors increasing that pair's relative likelihood, while a negative advantage favors decreasing it, subject to PPO clipping.
Reward credit therefore applies to the recorded continuation decisions, stop and action together.
The auxiliary value loss trains a separate prediction task on their shared representation.

\paragraph{Training objective.}
We train the joint policy with PPO, which compares the probability of each recorded decision under the current parameters with its probability under the frozen rollout parameters $\theta_{\rm old}$.
Because a decision is the pair $(a_t,\tau_t)$, the ratio $\rho_t(\theta)=p_\theta(a_t,\tau_t\mid s_t)/p_{\theta_{\rm old}}(a_t,\tau_t\mid s_t)$ includes the stopping probabilities.
Writing $\widehat{\E}_t$ for the average over rollout samples and $\widehat v_{\theta,d}(s)=V_{\theta,d}(z_{\theta,d}(s))$ for the value prediction at depth $d$ and holding rollout advantages and targets fixed, we minimize

\begin{align}
\cL(\theta)&=\cL_{\mathrm{PPO}}(\theta)
 +\frac{\lambda_V}{\dmax}\sum_{d=1}^{\dmax}\widehat{\E}_t
 \big[(\widehat v_{\theta,d}(s_t)-V_t^{\mathrm{targ}})^2\big]
 +\lambda_c\widehat{\E}_t\big[C_\theta(s_t)\big],\label{eq:full-loss}\\
C_\theta(s)&=\frac{1}{\dmax}\sum_{d=1}^{\dmax}d\,q_\theta(d\mid s)
 =\frac{1}{\dmax}\sum_{d=1}^{\dmax}S_{\theta,d}(s).\label{eq:compute-cost}
\end{align}

The cost gradient passes through the current stopping probabilities at the frozen rollout states.
This defines a local PPO regularizer; \Cref{app:training-math} distinguishes it from optimizing a trajectory return that includes future computation costs.

The cost term favors earlier stopping, while reward can favor continuing when the resulting action improves.

\paragraph{Shared value targets.}
Each intermediate representation is trained to predict the same realized GAE target~\citep{schulman2015high}, even if its action head was not selected.
The diagnostic

\begin{equation}
\widehat{\Voc}(s,d):=\widehat v_{\theta,d}(s)-\widehat v_{\theta,d-1}(s),
\qquad d=2,\ldots,\dmax,
\label{eq:voc}
\end{equation}

therefore compares predictions of a shared target.
The quantity $\widehat{\Voc}$ is a diagnostic of the value heads; halting is trained through the joint actor and cost gradients.
\Cref{prop:shared-target} characterizes the shared-target predictions and \Cref{app:calibration} shows why replacing their outputs leaves a frozen stopping policy unchanged.
Advantages use a state-only baseline that does not condition on the sampled depth; \Cref{alg:aic} and \Cref{app:training-math} give the full training procedure.


\subsection{Allocation Bound}
\label{sec:allocation-theory}
To study the opportunity for allocation, consider the return from choosing an action at each depth while holding future behavior fixed.
For depth-specific action distributions and a common continuation policy $\bar\pi$, define $W_d(s)=\E_{a\sim\pi_d(\cdot\mid s)}[Q^{\bar\pi}(s,a)]$.
An ideal allocator maximizes $W_d(s)-cd$.
Stopping at the first increment below cost is optimal when $W_{d+1}-W_d$ is non-increasing; otherwise a later gain can justify continuing through an unhelpful step (\Cref{app:halt-derivation}).
The matching cost units are $c=\lambda_c/\dmax$; this ideal problem is distinct from the PPO surrogate and shared-target critics.

The following result formalizes why varying depth across states can help at a common expected budget.

\begin{theorem}[Benefit of state-dependent allocation]
\label{thm:main}
Fix a state distribution $\mu$, a depth cap $\dmax\ge K\ge2$ and $0<\epsilon<1$. Easy states of mass $1-\epsilon$ attain reference value $W^*$ at every depth; hard states attain it at depths $d\ge K$ and have shortfall at least $\Delta(1-d/K)$ for $d<K$, where $\Delta>0$. An ideal allocator attains $W^*$ with expected depth $b=1+\epsilon(K-1)<K$. Every state-independent stopping distribution with $\E[\tau]\le b$ has expected shortfall at least
\begin{equation}
\mathcal R_\mu\ge\epsilon\Delta(1-b/K)>0.
\label{eq:allocation-lower-bound}
\end{equation}
\end{theorem}

The allocator reserves depth $K$ for hard states, while every state-independent schedule at the same expected budget has positive shortfall, including randomized schedules.
\Cref{app:proofs} gives the proof, extensions and sequential-return conditions.
The result concerns an ideal allocator that knows the easy/hard partition, with the state distribution and action policies held fixed.
Learning must recover a useful allocation from sampled outcomes.
For sequential tasks, \Cref{app:occupancy} relates the shortfall to return using the state distribution induced by each policy.

\section{Experiments}
\label{sec:matched}

Our experiments address three questions.
First, does AIC reach higher reward or lower computation than fixed-depth policies under the same cap on refinement steps (\Cref{sec:protocol})?
Second, do its gains come from where it spends computation, from the data it collects, or from value supervision (\Cref{sec:causal})?
Third, how does it compare with other learned halting methods (\Cref{sec:adaptive-baselines})?
We first describe the tasks, baselines and computation budgets shared by all three.

\paragraph{Tasks and baselines.}
The twelve-task benchmark covers continuous control, manipulation, navigation, language, gridworld and rare events (\Cref{app:tasks-app}).
GPT-2/GLUE-RTE is a single-step entailment-classification task (\Cref{app:tasks-llm}).
Fixed-depth baselines share AIC's encoder, refinement block, per-depth heads and PPO~\citep{schulman2017proximal} objective but always stop at a training depth $K_{\rm train}\in\{1,2,4,8\}$.
A privileged reference selects depth using simulator-side difficulty estimates $\hat\kappa(s)$ (\Cref{app:oracle}).
Hyperparameters, seeds and loss-weight sweeps appear in \Cref{app:hyperparams}.

\paragraph{Evaluation protocol.}
Benchmark results are five-seed means with standard deviations (\Cref{tab:full-budgets}); \Cref{tab:main} lists the $B=8$ means.
Each task keeps its own reward scale, so we compare methods within tasks and the POGP comparison uses POGP's return scale.
The data-source ablation, the stopping and supervision ablation and the memory, transfer and sensitivity studies use their own training runs and are not pooled with the benchmark, whereas the truncation ablation and the halting-baseline comparison reuse the benchmark checkpoints.

\paragraph{Computation budgets.}
The nominal budgets $B\in\{1,2,4,8\}$ cap executed depth: a fixed checkpoint uses $\min(K_{\rm train},B)$ refinements.
Fixed baselines come from a separate training sweep at each budget,
with learning rate and entropy coefficient tuned on held-out validation
episodes at that budget. For each seed, we select the checkpoint with
the highest validation return at budget $B$, evaluating each candidate
with $\min(K_{\mathrm{train}}, B)$ refinement steps.
Parentheses list the selected training depths, with seed counts when they differ (\Cref{app:matched-full}).
AIC at $B<8$ is retrained with $\dmax=B$ and the default hyperparameters, so at $B=1$ it has no stopping decision and the $B=1$ differences reflect training differences unrelated to allocation.
The main comparison places both methods under the same execution cap; \Cref{tab:aic-depth} gives AIC's realized depth at every budget and \Cref{tab:halt-step-app} every method's depth at $B=8$, while the sensitivity study in \Cref{app:hyperparams} matches mean depth instead.
\begin{table}
\centering
\caption{\textbf{Benchmark comparison at nominal $B{=}8$.}
$\Delta$: AIC minus the per-budget fixed baseline, whose selected training depths appear in parentheses with seed counts as superscripts. Gap recovery is measured toward the privileged reference and is undefined when the reference does not exceed the fixed baseline.
Best fixed: strongest fixed baseline at any budget $B'\le8$ in \Cref{tab:full-budgets}, with $B'$ in parentheses; $\Delta_{\rm best}$ is AIC minus that value.
$^{*}$: $p<.05$, two-sided Welch's $t$-test over five seeds with the standard deviations of \Cref{tab:full-budgets}.}
\label{tab:main}
\scriptsize
\setlength{\tabcolsep}{1.8pt}
\begin{tabular}{@{}l c c c c c c c c@{}}
\toprule
Task & Fixed ($K_{\rm train}$) & AIC & Privileged & $\Delta$ & \%$\Delta$ & Gap recovery & Best fixed ($B'$) & $\Delta_{\rm best}$ \\
\midrule
\TaskMGDist\,\scriptsize(sat.\ ctrl.) & $1.0000$ ($1$) & $1.0000$ & $1.0000$ & $.0000$ & $0.00\%$ & --- & $1.0000$ ($1$) & $.0000$ \\
\rowcolor{oursrow}\cellcolor{white}\TaskControl & $.9329$ ($8$)   & $.9348$  & $.9744$ & $+.0019^{*}$ & $+0.20\%$ & $4.6\%$ & $.9329$ ($8$) & $+.0019^{*}$ \\
\rowcolor{oursrow}\cellcolor{white}\TaskLLM     & $.8311$ ($8$)   & $.8334$  & $.9323$ & $+.0023$ & $+0.28\%$ & $2.3\%$ & $.8311$ ($8$) & $+.0023$ \\
\rowcolor{oursrow}\cellcolor{white}\TaskOGCube  & $.4281$ ($8$)   & $.4299$  & $.5646$ & $+.0018$ & $+0.42\%$ & $1.3\%$ & $.4281$ ($8$) & $+.0018$ \\
\rowcolor{oursrow}\cellcolor{white}\TaskNavNeg  & $.6704$ ($8$)   & $.6776$  & $.8685$ & $+.0072^{*}$ & $+1.07\%$ & $3.6\%$ & $.6916$ ($1$) & $-.0140^{*}$ \\
\rowcolor{oursrow}\cellcolor{white}\TaskOGScene & $.4075$ ($8$)   & $.4167$  & $.5565$ & $+.0092^{*}$ & $+2.26\%$ & $6.2\%$ & $.4075$ ($8$) & $+.0092^{*}$ \\
\rowcolor{oursrow}\cellcolor{white}\TaskStack   & $.4458$ ($8$)   & $.4579$  & $.6274$ & $+.0121^{*}$ & $+2.71\%$ & $6.7\%$ & $.4508$ ($4$) & $+.0071$ \\
\rowcolor{oursrow}\cellcolor{white}\TaskManip   & $.4741$ ($8$)   & $.4926$  & $.6756$ & $+.0185^{*}$ & $+3.90\%$ & $9.2\%$ & $.4911$ ($4$) & $+.0015$ \\
\rowcolor{oursrow}\cellcolor{white}\TaskNav     & $.5018$ ($2^{(1)},8^{(4)}$) & $.5209$  & $.7367$ & $+.0191^{*}$ & $+3.81\%$ & $8.1\%$ & $.5095$ ($4$) & $+.0114^{*}$ \\
\rowcolor{oursrow}\cellcolor{white}\TaskRare    & $.5575$ ($1^{(2)},8^{(3)}$) & $.5918$  & $.7523$ & $+.0343^{*}$ & $+6.15\%^{\dagger}$ & $17.6\%^{\dagger}$ & $.6338$ ($1$) & $-.0420^{*}$ \\
\midrule
\TaskOGPuzzle & $.4069$ ($8$) & $.4022$ & $.5185$ & $-.0047$ & $-1.16\%$ & $-4.2\%$ & $.4069$ ($8$) & $-.0047$ \\
\TaskMGFour   & $.9730$ ($8$) & $.9514$ & $.9730$ & $-.0216^{*}$ & $-2.22\%$ & --- & $.9730$ ($8$) & $-.0216^{*}$ \\
\bottomrule
\end{tabular}
\\[2pt]
{\scriptsize $^{\dagger}$On \TaskRare{}, the stopping and supervision ablation attributes most of the gain to value supervision (\Cref{app:supervision-vs-allocation}).}
\end{table}

\begin{figure}[t]
\centering
\includegraphics[width=\linewidth]{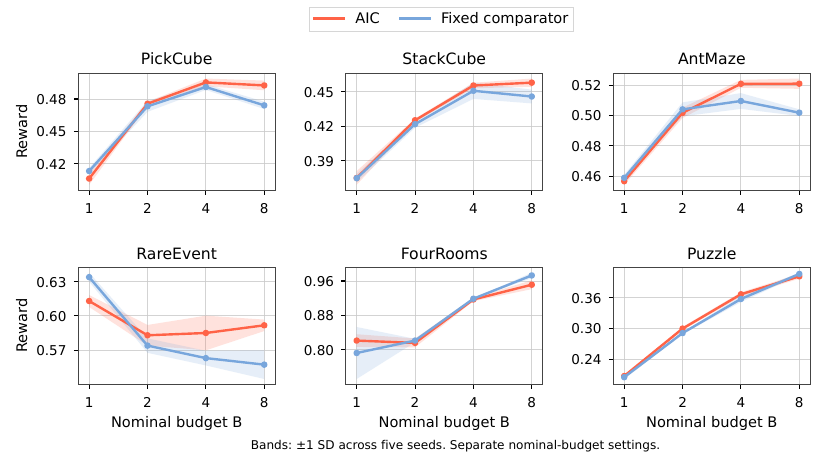}
\caption{\textbf{Performance across nominal budgets on six tasks.} Points are five-seed means with $\pm1$ standard-deviation bands (\Cref{tab:full-budgets}); each point is a separate budget setting and realized depth can differ between methods. RareEvent peaks at $B=1$ for both methods. \Cref{fig:pareto} shows the remaining six tasks.}
\label{fig:budget-main}
\end{figure}

\subsection{Comparison to Fixed-Depth Methods}
\label{sec:protocol}
A higher return can come from state-dependent stopping, intermediate value supervision, changes in the learned action policies, or different training trajectories.
We therefore ask what reward and computation the complete policy achieves relative to fixed and adaptive alternatives; how performance changes when we vary the data collector, the learner, the stopping configuration or the auxiliary supervision; and whether the frozen policy depends on refinement at the states where it selects greater depth.
The benchmark answers the first question, the data-source ablation and the stopping and supervision ablation answer the second and the truncation ablation answers the third, with an equal-work control that separates which states are truncated from how much computation is removed.

\paragraph{Fixed-depth results.}
At $B=8$, AIC exceeds the per-budget fixed baseline on nine tasks, seven of them at $p<.05$ (two-sided Welch's $t$-test over five seeds), ties on DistShift and trails on FourRooms and Puzzle (\Cref{tab:main}).
Against the strongest fixed baseline at any budget within the cap, AIC leads on seven tasks, three at $p<.05$, ties on one and trails on four: depth-one baselines lead on RareEvent ($.6338$) and low-heterogeneity AntMaze ($.6916$) and the PickCube margin narrows to $+.0015$ against Fixed-4, whose depth is closest to AIC's $3.47$.
Each budget has its own tuned sweep, so a per-budget baseline can trail another budget's: on PickCube, the $B=8$ sweep selects $K_{\rm train}=8$ ($.4741$), while the separately tuned $B=4$ sweep's Fixed-4 reaches $.4911$.
The largest per-budget gain is on RareEvent, where the stopping and supervision ablation attributes most of the change to value supervision and the privileged reference measures the remaining headroom relative to task-specific depth heuristics (\Cref{app:oracle}).

\paragraph{Comparison with POGP.}
\Cref{tab:pogp-aic} compares AIC with POGP on four MuJoCo locomotion tasks under an eight-step cap.
AIC uses 22--28\% fewer iterations per action on all four, while its mean return stays within 3--5 of POGP's, less than $0.3$ standard deviations.
Both methods use POGP's diffusion backbone, so one iteration is one denoising step in each (\Cref{app:pogp}).

\begin{table}
\centering
\caption{\textbf{POGP--AIC comparison with an eight-step cap.}
Undiscounted return over 100-step episodes (\Cref{app:pogp}) and iterations per action, mean $\pm$ standard deviation over five seeds.}
\label{tab:pogp-aic}
\small
\setlength{\tabcolsep}{8pt}
\begin{tabular}{@{}lcccc@{}}
\toprule
& \multicolumn{2}{c}{POGP} & \multicolumn{2}{c}{AIC} \\
\cmidrule(lr){2-3}\cmidrule(l){4-5}
Environment & Return $\uparrow$ & Steps $\downarrow$ & Return $\uparrow$ & Steps $\downarrow$ \\
\midrule
HalfCheetah & $330 \pm 22$ & $5.8\pm.3$ & $325 \pm 25$ & $4.2\pm.3$ \\
Walker2d & $180 \pm 15$ & $6.2\pm.3$ & $183 \pm 17$ & $4.8\pm.4$ \\
Ant & $91 \pm 10$ & $6.8\pm.4$ & $88 \pm 12$ & $5.3\pm.4$ \\
Hopper & $205 \pm 30$ & $5.9\pm.3$ & $208 \pm 28$ & $4.5\pm.3$ \\
\bottomrule
\end{tabular}
\end{table}

\subsection{Ablations and Controls}
\label{sec:causal}

AIC differs from a fixed-depth policy in three ways:
(1) it decides at each state how many refinement steps to run before acting (\Cref{sec:truncation});
(2) it collects different training data, because a policy that acts differently visits different states (\Cref{sec:yoked});
and (3) it trains a value head at every depth, which shapes the representation shared with the action and halting heads (\Cref{sec:factorial}).


\subsubsection{Truncation Study}
\label{sec:truncation}
We first ask whether the frozen policy relies more on refinement at the states where it chooses greater depth.
The targeted intervention forces high-depth decisions to stop at depth one and the control truncates an equally sized random subset, so a larger targeted drop indicates greater reliance on refinement at the selected states.
Matching the number of interventions does not match removed computation: truncating depth eight removes seven steps, while truncating depth two removes one.
For a reference set of decisions $\mathcal I$, forcing $A\subseteq\mathcal I$ to depth one removes $M(A)=\sum_{i\in A}(\tau_i-1)$ refinements, so sets of equal size can differ in $M(A)$.
We therefore also run an equal-work control (\Cref{app:critstate}).

\begin{table}
\centering
\caption{\textbf{Sensitivity to targeted and random truncation.}
Targeted: decisions at or above the 75th-percentile depth, forced to $d{=}1$. Random: count-matched random early exits.
Targeted drops exceed random drops on every non-saturated task, including the final two rows; ratios are omitted when the random drop is non-positive. \Cref{tab:equal-work} gives selected fractions, removed refinements and a random control matched on removed work.}
\label{tab:critical-ablation}
\scriptsize
\setlength{\tabcolsep}{4pt}
\begin{tabular}{@{}l c c c c c@{}}
\toprule
Task & Normal & Targeted Drop & Random Drop & Targeted$-$Random & Ratio \\
\midrule
\TaskMGDist\,\scriptsize(sat.) & $1.0000$ & $.0000$ & $.0000$ & $.0000$ & --- \\
\TaskNavNeg  & $.6776$ & $.0169$ & $-.0033$ & $.0202$ & --- \\
\TaskControl & $.9348$ & $.0247$ & $.0108$ & $.0139$ & $2.3\times$ \\
\TaskLLM     & $.8334$ & $.0501$ & $.0142$ & $.0359$ & $3.5\times$ \\
\TaskNav     & $.5209$ & $.0845$ & $.0228$ & $.0617$ & $3.7\times$ \\
\TaskManip   & $.4926$ & $.1062$ & $.0386$ & $.0676$ & $2.8\times$ \\
\TaskStack   & $.4579$ & $.1148$ & $.0446$ & $.0702$ & $2.6\times$ \\
\TaskOGCube  & $.4299$ & $.2091$ & $.1399$ & $.0692$ & $1.5\times$ \\
\TaskOGScene & $.4167$ & $.1945$ & $.1211$ & $.0734$ & $1.6\times$ \\
\TaskRare    & $.5918$ & $.0034$ & $-.0016$ & $.0050$ & --- \\
\midrule
\TaskMGFour  & $.9514$ & $.1189$ & $.0396$ & $.0793$ & $3.0\times$ \\
\TaskOGPuzzle& $.4022$ & $.2105$ & $.1553$ & $.0552$ & $1.4\times$ \\
\midrule
\multicolumn{2}{l}{\emph{mean over 9 positive comparisons}} & $.0894$ & $.0430$ & $.0463$ & --- \\
\bottomrule
\end{tabular}
\end{table}

Across the nine tasks where AIC beats its baseline, targeted truncation lowers reward by $0.0894$ on average and random truncation by $0.0430$, a difference of $0.0463$.
Part of this difference reflects removed work, because targeted decisions sit at greater depth: $4.80$ versus $2.47$ removed refinements per intervened decision on PickCube (\Cref{tab:equal-work}).
Matching removed refinements reduces the mean difference to $0.0118$, yet the targeted drop still exceeds the equal-work drop by $.0065$--$.0219$ on PickCube, StackCube, both AntMaze variants, RTE, RareEvent and FourRooms; on HalfCheetah, Cube, Scene and Puzzle the two agree within $.004$.
RareEvent shows small truncation effects, consistent with its large supervision effect in the stopping and supervision ablation.
FourRooms and Puzzle show positive intervention gaps despite trailing their fixed baselines, so reliance on refinement and improvement over a separately trained baseline are distinct properties of a policy.

\subsubsection{Data-Source Ablation}
\label{sec:yoked}

When training a policy with RL, a policy that acts differently will inevitably see different data. 
To separate the data from the policy, we swap them on \TaskManip{} and \TaskStack{}.
A Fixed-$1$ policy and AIC at $B=8$ are each trained, frozen and used to collect one fixed-size dataset used to train a new Fixed-$1$ and AIC learners from scratch through behavioral cloning and value distillation.


\begin{table}
\centering
\caption{\textbf{Data-source ablation.}
Rows are collector--learner pairs; columns give reward and the learner's mean halting depth $\E$[steps].
The combined gain exceeds the sum of the separate gains.
Values are means $\pm$ standard deviation over five learner seeds, each collector produced one dataset.}
\label{tab:yoked}
\scriptsize
\setlength{\tabcolsep}{6pt}
\begin{tabular}{@{}l cc cc@{}}
\toprule
& \multicolumn{2}{c}{\TaskManip} & \multicolumn{2}{c}{\TaskStack} \\
\cmidrule(lr){2-3}\cmidrule(lr){4-5}
Configuration & Reward & $\E[\text{steps}]$ & Reward & $\E[\text{steps}]$ \\
\midrule
Fixed-1 collector $\to$ Fixed-1 learner & $.3195 \pm .025$ & $1.00\pm.00$ & $.2854 \pm .030$ & $1.00\pm.00$ \\
Fixed-1 collector $\to$ AIC learner     & $.3565 \pm .018$ & $1.00\pm.00$ & $.3194 \pm .022$ & $1.00\pm.00$ \\
AIC collector $\to$ Fixed-1 learner     & $.3522 \pm .013$ & $1.00\pm.00$ & $.3241 \pm .014$ & $1.00\pm.00$ \\
\rowcolor{oursrow}AIC collector $\to$ AIC learner & $.4873 \pm .003$ & $3.34\pm.06$ & $.4431 \pm .002$ & $3.99\pm.08$ \\
\bottomrule
\end{tabular}
\end{table}

Changing only the collector or only the learner adds a reward of $0.03$--$0.04$, while changing both adds $0.16$--$0.17$ (\Cref{tab:yoked}).
The combined gain exceeds the sum of the two separate gains by $.0981$ on PickCube and $.0850$ on StackCube, with standard errors of $.015$ and $.018$.
AIC's gain therefore does not come from its training data alone, since most of it appears only when an AIC learner is trained on AIC data, the one setting that also refines beyond depth one.


\subsubsection{Stopping and Supervision Ablation}
\label{sec:factorial}
AIC adds two components to a one-step policy, adaptive stopping and value supervision at every depth, which we will isolate in this experiment.
We train four configurations independently, with each component switched on or off (\Cref{tab:alloc-vs-sup}).
Let $R_{as}$ denote reward with adaptive stopping $a\in\{0,1\}$ and auxiliary supervision $s\in\{0,1\}$.
The stopping effects $R_{10}-R_{00}$ and $R_{11}-R_{01}$ measure the reward change from enabling stopping without and with supervision.
The supervision effects $R_{01}-R_{00}$ and $R_{11}-R_{10}$ measure the change from enabling supervision without and with stopping.

\begin{table}
\centering
\caption{\textbf{Stopping and Supervision Ablation.} Each configuration gives reward mean $\pm$ standard deviation over five seeds and mean executed depth $\bar d$ for an independently trained configuration. The non-adaptive configurations execute a fixed schedule of one refinement per decision. the adaptive configurations use AIC's learned stopping with $\dmax=8$.}
\label{tab:alloc-vs-sup}
\scriptsize
\setlength{\tabcolsep}{3.2pt}
\begin{tabular}{@{}l cc cc cc cc@{}}
\toprule
& \multicolumn{2}{c}{Neither} & \multicolumn{2}{c}{Supervision only} & \multicolumn{2}{c}{Adaptive stopping only} & \multicolumn{2}{c}{Full AIC}\\
\cmidrule(lr){2-3}\cmidrule(lr){4-5}\cmidrule(lr){6-7}\cmidrule(l){8-9}
Task & Reward & $\bar d$ & Reward & $\bar d$ & Reward & $\bar d$ & Reward & $\bar d$\\
\midrule
PickCube & $.421\pm.006$ & 1.00 & $.434\pm.005$ & 1.00 & $.472\pm.004$ & 3.61 & $.488\pm.004$ & 3.52\\
AntMaze & $.451\pm.005$ & 1.00 & $.458\pm.004$ & 1.00 & $.505\pm.004$ & 3.28 & $.518\pm.003$ & 3.19\\
Memory & $.503\pm.007$ & 1.00 & $.515\pm.006$ & 1.00 & $.552\pm.006$ & 3.42 & $.568\pm.005$ & 3.33\\
DoorKey & $.486\pm.007$ & 1.00 & $.499\pm.006$ & 1.00 & $.536\pm.005$ & 3.49 & $.551\pm.005$ & 3.38\\
StackCube & $.394\pm.006$ & 1.00 & $.413\pm.006$ & 1.00 & $.441\pm.005$ & 3.94 & $.455\pm.004$ & 3.81\\
RareEvent & $.558\pm.008$ & 1.00 & $.589\pm.006$ & 1.00 & $.560\pm.009$ & 1.92 & $.591\pm.006$ & 1.84\\
\bottomrule
\end{tabular}
\end{table}
On PickCube the stopping effects are $.051$ and $.054$ and the supervision effects $.013$ and $.016$; on AntMaze they are $.054$ and $.060$ versus $.007$ and $.013$.
RareEvent reverses the pattern: stopping changes reward by $.002$ at either supervision setting, while supervision changes it by $.031$.
StackCube gains from both factors and the memory tasks Memory and DoorKey follow PickCube (\Cref{app:pomdp}).
Because the non-adaptive configurations execute one refinement per decision and the adaptive configurations execute 1.8--3.9, each stopping effect combines state-dependent allocation with additional computation.
The Fixed-4 baseline of \Cref{tab:full-budgets}, trained separately,
reaches $.4911$ on PickCube and $.5095$ on AntMaze, above the
corresponding adaptive-stopping-only configurations.
Appendix~\ref{app:supervision-vs-allocation} outlines a comparison
at matched average computation that could isolate the effect of allocation.

\subsection{Baseline Comparisons with Other Halting Methods}
\label{sec:adaptive-baselines}

We next compare AIC with other stopping methods on seven tasks: four untrained rules (random stopping, confidence thresholding, action agreement and a value-gain threshold) and two jointly trained methods, ACT-ponder~\citep{graves2016adaptive} and a learned non-VoC threshold.
\Cref{app:baseline-defs} defines all methods and \Cref{tab:halt-step-app} gives their mean depths.

\begin{table}
\centering
\caption{\textbf{Adaptive halting baselines at $B{=}8$ across 7 tasks.}
Mean reward over five seeds, with the seed standard deviation as a subscript; AIC and Fixed repeat \Cref{tab:full-budgets}.
The trained methods reach similar rewards on several tasks at different mean depths.
ACT-ponder leads on FourRooms and the random schedule, which executes one refinement there, leads on RareEvent.}
\label{tab:baselines-adaptive}
\scriptsize
\setlength{\tabcolsep}{1.6pt}
\begin{tabular}{@{}l ccc cccc c@{}}
\toprule
& \multicolumn{3}{c}{Trained jointly with the policy} & \multicolumn{4}{c}{Untrained stopping rules} & \\
\cmidrule(lr){2-4}\cmidrule(lr){5-8}
Task & AIC & ACT-ponder & Lrn non-VoC & Conf.\ thr. & Action-agr. & Val.-gain thr. & Random & Fixed \\
\midrule
PickCube & $.4926_{\pm.0045}$ & $.4910_{\pm.0041}$ & $.4907_{\pm.0047}$ & $.4523_{\pm.0062}$ & $.4589_{\pm.0057}$ & $.4425_{\pm.0066}$ & $.4564_{\pm.0049}$ & $.4741_{\pm.0034}$ \\
StackCube & $.4579_{\pm.0036}$ & $.4521_{\pm.0039}$ & $.4527_{\pm.0042}$ & $.4218_{\pm.0058}$ & $.4378_{\pm.0061}$ & $.4066_{\pm.0072}$ & $.4163_{\pm.0053}$ & $.4458_{\pm.0060}$ \\
AntMaze (main) & $.5209_{\pm.0034}$ & $.5200_{\pm.0031}$ & $.5208_{\pm.0036}$ & $.4626_{\pm.0049}$ & $.4972_{\pm.0044}$ & $.4680_{\pm.0058}$ & $.4824_{\pm.0047}$ & $.5018_{\pm.0022}$ \\
GLUE-RTE & $.8334_{\pm.0009}$ & $.8300_{\pm.0042}$ & $.8299_{\pm.0040}$ & $.7617_{\pm.0031}$ & $.8110_{\pm.0026}$ & $.7957_{\pm.0029}$ & $.7932_{\pm.0034}$ & $.8311_{\pm.0024}$ \\
Scene & $.4167_{\pm.0048}$ & $.4130_{\pm.0044}$ & $.4136_{\pm.0051}$ & $.3884_{\pm.0055}$ & $.3530_{\pm.0063}$ & $.3716_{\pm.0059}$ & $.3903_{\pm.0052}$ & $.4075_{\pm.0044}$ \\
\midrule
RareEvent & $.5918_{\pm.0050}$ & $.5868_{\pm.0058}$ & $.5870_{\pm.0055}$ & $.5797_{\pm.0071}$ & $.4198_{\pm.0098}$ & $.4198_{\pm.0098}$ & $.6178_{\pm.0046}$ & $.5575_{\pm.0125}$ \\
FourRooms & $.9514_{\pm.0103}$ & $.9658_{\pm.0087}$ & $.9586_{\pm.0096}$ & $.8829_{\pm.0134}$ & $.7748_{\pm.0182}$ & $.7315_{\pm.0191}$ & $.7045_{\pm.0213}$ & $.9730_{\pm.0064}$ \\
\bottomrule
\end{tabular}
\end{table}

\begin{figure}[t]
\centering
\includegraphics[width=0.8\linewidth]{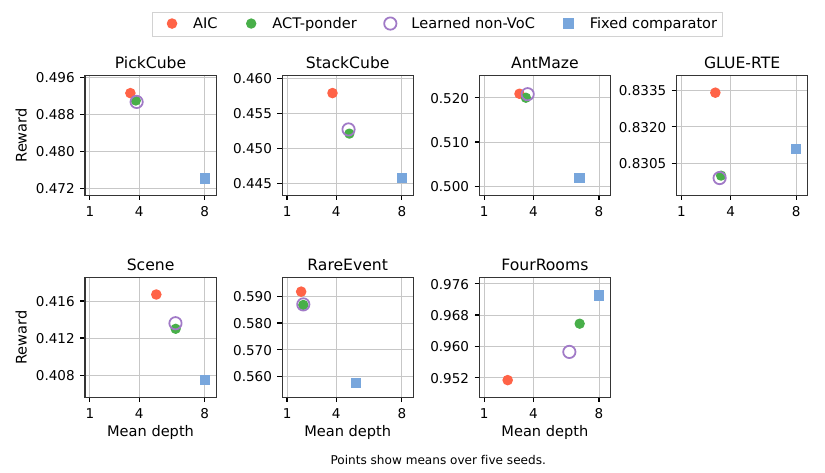}
\caption{\textbf{Reward versus mean halting depth at nominal $B=8$.} Panels show AIC, ACT-ponder, learned non-VoC and fixed baselines, using \Cref{tab:baselines-adaptive,tab:halt-step-app}. A fixed baseline's depth is the seed average of $\min(K_{\rm train},8)$ over its selected checkpoints.}
\label{fig:depth-bars}
\end{figure}

AIC's reward is close to that of the trained baselines, exceeding ACT-ponder by $.0016$ on PickCube and learned non-VoC by $.0001$ on AntMaze, both within one standard error of the difference.
Across the seven tasks, AIC is within two standard errors of both trained baselines except on StackCube, where AIC is higher and FourRooms, where ACT-ponder is higher.
AIC uses the lowest mean depth of the three on all seven tasks (\Cref{fig:depth-bars}); on FourRooms that lower depth comes with lower reward.

The learned non-VoC baseline shares AIC's encoder, refinement block, action heads and value heads and differs only in how its halting head is trained: its stopping probabilities neither enter the PPO ratio nor receive advantage credit and the head instead learns from a stop-gradient copy of $z_d$ to predict whether the next refinement leaves the action distribution unchanged.
The comparison therefore separates return-based stopping credit from a learned convergence signal on the same representation (\Cref{tab:halt-spec}).
Parameter counts are $2.071$M for AIC, $2.069$M for ACT-ponder and $2.110$M for learned non-VoC.
On GPT-2/RTE, with a checkpoint trained at $\lambda_c=.003$ and mean depth $4.08$, AIC executes 49\% fewer refinements than Fixed-8 and lowers end-to-end latency by 7.8\%, since the frozen encoder dominates the total cost.
Training evaluates every depth for value supervision. On PickCube, AIC training took $1.22\times$ as long as Fixed-8 training (\Cref{app:efficiency}).
Overall, training the halting head through the return matches or exceeds the reward of the other learned halting methods on six of seven tasks while executing fewer refinement steps on all seven.

\section{Discussion}
\label{sec:failure-modes}

Our analysis identifies three ingredients for useful allocation: state-dependent benefits from refinement, performance headroom and a representation from which the allocator can recognize those benefits.
The tasks differ in where the gains come from.
On PickCube and AntMaze, adaptive stopping, which also executes more refinements, contributes more than supervision; on RareEvent, supervision contributes more.

FourRooms is the clearest failure: AIC trails its mean-depth-matched baseline at all twenty loss-weight settings, by $0.7$ to $2.1$ percentage points of normalized reward (\Cref{tab:fourrooms-sweep}; best $-.7\pm.3$ at $(\lambda_c,\lambda_V)=(.001,.5)$).
Puzzle, in contrast, turns positive with stronger value supervision (\Cref{tab:puzzle-sweep}), so its default-setting loss depends on the training configuration.
Stopping and the action policies at intermediate depths must therefore be evaluated together.

Attached to DiffusionQL, DPPO, D3P, DSAC-D and IDQL (\Cref{tab:base_policy_task_gain}), AIC raises mean return on six of eight tasks in all five families, lowers it on FourRooms in all five and changes sign across families on Puzzle. HalfCheetah gains lie within one standard error in every family, gains on the other five tasks exceed two standard errors except DSAC-D on Scene and every FourRooms decrease exceeds two standard errors.
\Cref{app:transfer} defines each family's refinement step and intermediate-action decoder and \Cref{app:pomdp} extends AIC to memory tasks.

\begin{table}
\centering
\caption{\textbf{Transfer to five diffusion-based policy families at nominal $B=8$.}
Gain $\Delta=R_{\mathrm{AIC}}-R_{\mathrm{Fixed}}$ with its standard error over five seeds as a subscript. Fixed executes eight refinement steps per action in every family; \Cref{tab:transfer-full} lists the returns and AIC's mean halting depth.}
\label{tab:base_policy_task_gain}
\scriptsize
\setlength{\tabcolsep}{4pt}
\begin{tabular}{@{}l ccccc@{}}
\toprule
Task & DiffusionQL & DPPO & D3P & DSAC-D & IDQL \\
\midrule
HalfCheetah-v4 & $+.002_{\pm.003}$ & $+.001_{\pm.002}$ & $+.002_{\pm.003}$ & $+.001_{\pm.003}$ & $+.001_{\pm.002}$ \\
ManiSkill PickCube-v1 & $+.020_{\pm.003}$ & $+.018_{\pm.003}$ & $+.023_{\pm.003}$ & $+.010_{\pm.004}$ & $+.012_{\pm.003}$ \\
ManiSkill StackCube-v1 & $+.015_{\pm.003}$ & $+.013_{\pm.003}$ & $+.018_{\pm.003}$ & $+.008_{\pm.003}$ & $+.010_{\pm.003}$ \\
OGBench scene-play-v0 & $+.012_{\pm.003}$ & $+.008_{\pm.003}$ & $+.014_{\pm.003}$ & $+.004_{\pm.003}$ & $+.010_{\pm.003}$ \\
AntMaze (main) & $+.017_{\pm.003}$ & $+.015_{\pm.002}$ & $+.018_{\pm.003}$ & $+.010_{\pm.003}$ & $+.014_{\pm.003}$ \\
OGBench puzzle-4x4-play-v0 & $-.003_{\pm.003}$ & $-.005_{\pm.003}$ & $+.002_{\pm.003}$ & $-.006_{\pm.002}$ & $+.001_{\pm.003}$ \\
MiniGrid-FourRooms & $-.015_{\pm.005}$ & $-.018_{\pm.005}$ & $-.012_{\pm.004}$ & $-.020_{\pm.004}$ & $-.014_{\pm.004}$ \\
RareEvent-StackCube-Misalign & $+.025_{\pm.005}$ & $+.022_{\pm.005}$ & $+.030_{\pm.005}$ & $+.012_{\pm.005}$ & $+.018_{\pm.005}$ \\
\bottomrule
\end{tabular}
\end{table}


\paragraph{Limitations and future work.}
While our controls cannot fully separate where computation is spent from how much is spent, they show that the policy depends more on refinement at the states where it chooses greater depth, a difference that persists on seven tasks when the removed computation is matched.
Reassigning recorded depths across states at fixed total computation would isolate this state--depth association further.
Adaptive stopping does not help on every task and FourRooms remains below a fixed depth across the loss weights we tried on eleven of twelve tasks. however, against the fixed baseline tuned at the same eight-step budget, AIC had higher mean reward on nine tasks, tied on one, and had lower mean reward on two, while averaging 3.3 refinement steps.
Our experiments use small refinement modules in simulation, yet the gains transfer to five diffusion-policy families, which suggests the approach is not tied to one architecture.
For future work, we plan on scaling AIC to large backbones and real robots.

\section{Conclusion}
\label{sec:conclusion}
We introduced AIC, an iterative policy that treats stopping depth and action as one decision, so that environment return trains both.
We showed that when only some states benefit from further refinement, an ideal state-dependent policy outperforms every state-independent schedule at the same expected depth.
On twelve benchmark tasks with at most eight refinement steps per decision, AIC executes 3.3 steps on average; its mean reward exceeds the best fixed-depth policy on seven tasks and trails it on four.
It also uses fewer steps than two jointly trained halting methods on all seven compared tasks and than POGP on four locomotion tasks, at comparable reward.
Ablating adaptive stopping and value supervision separately shows that stopping accounts for most of the reward gain on PickCube and AntMaze and supervision on RareEvent.
The adaptive runs also execute more refinement steps. Isolating the
effect of allocation would require the work-matched comparison
outlined in Appendix~\ref{app:supervision-vs-allocation}.
Learning when to think thus becomes part of learning how to act.

\paragraph{Acknowledgements}
The authors acknowledge Advanced Research Computing at Virginia Tech for providing computational resources and technical support that have contributed to the results reported within this paper. URL: \url{https://arc.vt.edu/}

\clearpage

{\small
\bibliographystyle{abbrvnat}
\bibliography{ref}
}

\clearpage
\appendix
\renewcommand{\thesection}{\Alph{section}}

\section*{Supplementary Material}

The supplement begins with the method and its derivations, followed by the allocation theorem and the ideal stopping problem.
\Cref{app:arch} follows one decision from encoding to action, derives stopping probabilities and the PPO update and explains the role of each output head.
\Cref{app:proofs} gives a concrete allocation example and a step-by-step proof of \Cref{thm:main}, followed by its extensions and sequential-return conditions.
\Cref{app:halt-derivation} derives optimal stopping for a known depth-value profile and identifies when a one-step threshold is sufficient.

\Cref{app:tasks-app} then defines the tasks.
The empirical sections provide the prior-work comparison, factorial interpretation, adaptive baselines, computation accounting and sensitivity studies (\Cref{app:precursor,app:supervision-vs-allocation,app:adaptive-baselines-extra,app:halt-step-full,app:hyperparams}).
The remaining sections develop the privileged reference, full budget results, intervention and annotation procedures, allocation profiles and task-level interpretation (\Cref{app:oracle,app:matched-full,app:critstate,app:pr-full,app:regime-detail}).
\Cref{app:limitations,app:pomdp,app:efficiency,app:transfer,app:inventory} collect the limitations, memory extensions, timing measurements, transfer details and a protocol summary.

\section{Method Details}
\label{app:arch}
The central distinction is between an environment step, which produces a new observation and reward and a refinement step, which processes the current observation again.

\subsection{Environment steps and internal computation}
\label{app:method-walkthrough}
At step $t$, the state is $s_t$, the action is $a_t$, the reward is $r_t$ and the next state is $s_{t+1}$.
The policy's parameters, denoted by $\theta$, are the neural-network weights that training adjusts.

The discounted return from step $t$ is the sum $G_t=\sum_{k\ge0}\gamma^k r_{t+k}$, stopped at the end of a finite episode when applicable.
For continuing tasks with bounded rewards, $0\le\gamma<1$ ensures a finite return; bounded-horizon episodes can also use $\gamma=1$.
A value prediction estimates this return.

AIC adds a second index, $d$, for computation within one environment step.
The policy encodes $s_t$, refines that encoding and decides whether to continue refining before it sends an action to the environment.
The state, reward clock and available observation remain fixed throughout this internal sequence.
Thus $d=3$ means three refinements of the current decision, not three environment transitions.
In this formulation, discounting applies across environment steps and the separate depth penalty charges for internal computation.

For a fully observed Markov decision process, $s_t$ contains the information needed to predict the next-state distribution given an action.
In a partially observed task, the policy instead needs a suitable representation of its observation history.
The notation $s_t$ can represent the information supplied to that policy, including its history representation.
The memory-task description in \Cref{app:pomdp} therefore distinguishes observation history from internal refinement.

\subsection{Encoding, refinement and the three outputs}
\label{app:forward-walkthrough}
An encoder converts the input into a vector of numbers, $z_0=f_{\rm enc}(s)$.
The encoder depends on the input type.
A multilayer perceptron (MLP) processes proprioceptive inputs.
A convolutional neural network (CNN) processes grid inputs and the language task uses the frozen GPT-2 representation described in \Cref{app:tasks-llm}.

The refinement block repeatedly transforms the encoded vector:
\[
z_d=z_{d-1}+F_\theta(z_{d-1}),\qquad d=1,\ldots,\dmax.
\]
Here $F_\theta$ is the shared MLP block.
The same block parameters are used at every depth, so increasing depth repeats computation without allocating a different refinement network to each iteration.
All representations describe the current decision; their action quality can vary non-monotonically with depth.

Three lightweight heads read the representation at a depth.
The action head $\pi_{\theta,d}$ gives the distribution used if that depth is selected.
For a discrete action space it assigns probabilities to actions; for a continuous action space its likelihood is a probability density.
The value head $V_{\theta,d}$ produces a scalar prediction of the shared training target.
The halting head produces $h_{\theta,d}\in[0,1]$, the probability of stopping if depth $d$ has been reached.
Only depths below $\dmax$ require a learned stopping probability, because stopping at the cap is compulsory.

The main configuration uses latent dimension 256, cap eight and depth-specific heads.
Throughout the appendix, $\pi_{\theta,d}(a\mid s)$ abbreviates $\pi_{\theta,d}(a\mid z_{\theta,d}(s))$ and $\widehat v_{\theta,d}(s)=V_{\theta,d}(z_{\theta,d}(s))$.
The hats distinguish learned value predictions from the depth-specific returns used in the allocation analysis.

\subsection{Stopping-distribution example}
\label{app:stopping-example}
Consider one state with a cap of four solely to make the calculation short.
Suppose its three conditional stopping probabilities are $h_1=.2$, $h_2=.6$ and $h_3=.25$.
At depth one, the policy stops with probability $.2$ and continues with probability $.8$.
The probability of stopping at depth two is therefore $.8\times.6=.48$, not $.6$: the policy must first reach that depth.
Continuing through the first two depths has probability $.8\times.4=.32$.
Of that remaining mass, $.32\times.25=.08$ stops at depth three and $.32\times.75=.24$ reaches the forced cap.
Consequently,
\[
(S_1,S_2,S_3,S_4)=(1,.8,.32,.24),\qquad
(q_1,q_2,q_3,q_4)=(.2,.48,.08,.24).
\]
The entries of $q$ sum to one.
The survival probability $S_d$ answers whether depth $d$ is reached; $q_d$ answers whether it is the final depth.
Conditional stopping probabilities $h_d$ need not sum to one, because each applies to a different conditional event.

The expected depth can be calculated in either of two equivalent ways:
\[
\E[\tau\mid s]
=1(.2)+2(.48)+3(.08)+4(.24)
=1+.8+.32+.24=2.36.
\]
The first expression weights each final depth by its probability.
The second counts one unit of computation for every depth reached: the first is always reached, the second with probability $.8$ and so on.
Normalizing by the cap gives $C(s)=2.36/4=.59$.
The value $.59$ expresses expected depth as a fraction of the cap.

Suppose the sampled depth is two and a particular discrete action has probability $.7$ under $\pi_2$.
The joint probability of that depth and action is $.48\times.7=.336$.
The stored log-probability is $\log(.8)+\log(.6)+\log(.7)$, which records continuing, stopping and choosing the action.
At the cap, the stopping factor is one and contributes zero log-probability.
Although the environment receives only the action, the training record retains the depth that produced it.

\subsection{Stopping probabilities and expected computation}
\label{app:stopping-law}
Expected depth can be computed by adding the probabilities of reaching each refinement step. This gives a differentiable computation penalty without sampling a depth.
For a fixed state, abbreviate $h_d=h_{\theta,d}(s)$ and $S_d=\prod_{j<d}(1-h_j)$. Thus $S_1=1$ and $S_d$ is the probability of reaching depth $d$. The cap is forced: only $h_1,\ldots,h_{\dmax-1}$ are free stopping probabilities.

\begin{lemma}[Stopping-law identities]
\label{lem:stopping-law}
For any $h\in[0,1]^{\dmax-1}$, the distribution in \Cref{eq:joint-policy} is normalized, with
\begin{equation}
\Pr(\tau\ge d\mid s)=S_d,\qquad
\E[\tau\mid s]=\sum_{d=1}^{\dmax}S_d.
\label{eq:survival-identities}
\end{equation}
Its normalized mean $C(s)=\E[\tau\mid s]/\dmax$ satisfies, for $j<\dmax$,
\begin{equation}
\frac{\partial C(s)}{\partial h_j}
=-\frac{S_j}{\dmax}\sum_{k=j+1}^{\dmax}
\prod_{\ell=j+1}^{k-1}(1-h_\ell)\le0.
\label{eq:compute-hazard-gradient}
\end{equation}
\end{lemma}
\begin{proof}
\emph{Step 1: account for every possible final depth.}
Reaching depth $d$ has probability $S_d$.
Of that mass, $S_dh_d$ stops and $S_d(1-h_d)=S_{d+1}$ continues.
Thus, below the cap,
\[
q(d)=S_dh_d=S_d-S_{d+1},
\qquad q(\dmax)=S_{\dmax}.
\]
The last equality assigns all remaining probability to the forced stop.
Adding these nonnegative masses cancels each intermediate survival term:
\[
\sum_{d=1}^{\dmax}q(d)
=(S_1-S_2)+(S_2-S_3)+\cdots+(S_{\dmax-1}-S_{\dmax})+S_{\dmax}
=S_1=1.
\]
The same cancellation starting at $d$ gives
$\sum_{k=d}^{\dmax}q(k)=S_d$, which is exactly $\Pr(\tau\ge d\mid s)$.
For $\dmax=1$, the intermediate sums are empty and $q(1)=S_1=1$.

\emph{Step 2: count executed steps instead of final depths.}
For a realized stop at $\tau$, the indicator $\mathbf 1\{\tau\ge d\}$ is one for each executed refinement and zero for every unexecuted refinement.
Consequently,
\[
\tau=\sum_{d=1}^{\dmax}\mathbf 1\{\tau\ge d\},
\qquad
\E[\tau\mid s]=\sum_{d=1}^{\dmax}\Pr(\tau\ge d\mid s)
=\sum_{d=1}^{\dmax}S_d.
\]
This explains the two equivalent computations of $2.36$ in \Cref{app:stopping-example}.

\emph{Step 3: differentiate the cost of the steps that can be avoided.}
Changing $h_j$ cannot affect whether an earlier depth $k\le j$ was reached.
For a later depth $k>j$, factor its survival probability as
\[
S_k=S_j(1-h_j)\prod_{\ell=j+1}^{k-1}(1-h_\ell).
\]
Here $S_j$ uses only earlier hazards and the product uses only later ones.
Differentiating the finite sum from Step 2 therefore gives
\[
\frac{\partial C}{\partial h_j}
=\frac{1}{\dmax}\sum_{k=j+1}^{\dmax}\frac{\partial S_k}{\partial h_j}
=-\frac{S_j}{\dmax}\sum_{k=j+1}^{\dmax}
\prod_{\ell=j+1}^{k-1}(1-h_\ell).
\]
Every factor is nonnegative, so increasing a stopping probability cannot increase expected depth when the other hazards are held fixed.
An empty product equals one, including the term $k=j+1$.
The expression is a polynomial, so it remains valid at $h_j=1$ without dividing by $1-h_j$.
\end{proof}

Equation~\eqref{eq:compute-hazard-gradient} is a partial derivative in hazard coordinates; shared network parameters affect multiple hazards and representations, so the full parameter derivative uses the chain rule. For differentiable positive $q_\theta$, the equivalent score identity is
\begin{equation}
\nabla_\theta C_\theta(s)=
\E_{\tau\sim q_\theta}\!\left[\frac{\tau}{\dmax}\nabla_\theta\log q_\theta(\tau\mid s)\right].
\label{eq:cost-score}
\end{equation}
To obtain this identity, differentiate the finite expectation and use
$\nabla_\theta q_\theta(d\mid s)=q_\theta(d\mid s)\nabla_\theta\log q_\theta(d\mid s)$:
\[
\nabla_\theta C_\theta(s)
=\sum_d\frac{d}{\dmax}\nabla_\theta q_\theta(d\mid s)
=\sum_dq_\theta(d\mid s)\frac{d}{\dmax}\nabla_\theta\log q_\theta(d\mid s).
\]
A state-only baseline can be subtracted while holding it fixed, since $\sum_d\nabla q_\theta(d\mid s)=0$.
The analytic expression in \Cref{eq:compute-cost} avoids sampling altogether.
If a score estimator reuses old-policy depth samples, it needs importance weights for the current $q_\theta$; differentiating a detached integer depth does not supply this gradient.

\subsection{Returns, advantages and the shared value target}
\label{app:target-walkthrough}
An observed return alone does not say whether a decision was good relative to what was already expected in that state.
A state-only baseline $b_{\rm old}(s_t)$ supplies that comparison.
It is computed under the frozen rollout convention and does not condition on the sampled depth or action.
An advantage is a signed estimate of the difference between an outcome and this baseline.
A positive advantage favors the recorded joint decision; a negative advantage disfavors it.
Each sampled outcome supplies credit to the recorded depth and action.

Generalized advantage estimation (GAE) combines successive temporal-difference residuals.
A residual compares the current value prediction with one reward plus the discounted prediction at the next state:
\[
\delta_t=r_t+\gamma m_t b_{\rm old}(s_{t+1})-b_{\rm old}(s_t).
\]
The mask $m_t$ is zero after a true terminal transition, because no continuation value remains and one otherwise.
For a finite rollout segment, the advantages can be accumulated backward as
\[
\widehat A_t=\delta_t+\gamma\lambda_{\rm GAE}m_t\widehat A_{t+1},
\qquad \widehat A_T=0.
\]
This recursion is the finite sum in \Cref{eq:gae-shared}.
The coefficient $\lambda_{\rm GAE}$ controls how much later residuals contribute; it is distinct from both loss weights $\lambda_V$ and $\lambda_c$.
At a nonterminal rollout cutoff, the last residual still includes the bootstrap value of the next state.
Setting $\widehat A_T=0$ ends the residual accumulation and does not remove that bootstrap.

For a worked calculation, take a two-transition segment with rewards $.1$ and $.7$, with the second transition terminal.
Let the baseline predictions at the two visited states be $.2$ and $.3$ and use $\gamma=.9$ and $\lambda_{\rm GAE}=.8$ for this calculation.
Then
\[
\begin{aligned}
\delta_t&=.1+.9(.3)-.2=.17, & \delta_{t+1}&=.7-.3=.4,\\
\widehat A_t&=.17+.9(.8)(.4)=.458, &
V_t^{\rm targ}&=.2+.458=.658.
\end{aligned}
\]
Every intermediate critic at state $s_t$ receives the same target $.658$.
The target is a bootstrapped return estimate; it need not equal the Monte Carlo return, which here is $.1+.9(.7)=.73$.
The example also shows why the target is not the observed outcome of every possible stopping depth: only the sampled depth and its action were executed.

The objective uses a state-only baseline and frozen targets.
During a PPO update, the old baseline, advantage and value target are constants with respect to the current parameters.
\Cref{app:training-math} specifies the terminal and rollout-cutoff conventions.

\subsection{Joint credit under PPO}
\label{app:ppo-walkthrough}
The sampled pair $(\tau_t,a_t)$ is treated as one policy decision.
Its log-probability separates into a depth term and an action term:
\[
\log p_\theta(a_t,\tau_t\mid s_t)
=\log q_\theta(\tau_t\mid s_t)
 +\log\pi_{\theta,\tau_t}(a_t\mid s_t).
\]
The depth term includes each decision to continue before stopping.
Before clipping and regularization, the likelihood-ratio actor signal is proportional to this log-probability's gradient multiplied by the advantage.
It can therefore train stopping from task outcomes without a supervised label for the correct depth.
Using only the action likelihood would omit the direct score term for the recorded stopping path.

For a sampled stop at depth two below the cap, $\log q_2=\log(1-h_1)+\log h_2$.
Holding the other outputs fixed and taking interior probabilities,
\[
\frac{\partial\log q_2}{\partial h_1}=-\frac{1}{1-h_1},\qquad
\frac{\partial\log q_2}{\partial h_2}=\frac{1}{h_2}.
\]
A positive advantage thus favors continuing past depth one and stopping at depth two for this sampled path.
A negative advantage reverses these score directions.
These are derivatives with respect to the output probabilities; shared network parameters couple the outputs and the final update also includes clipping, value regression and the computation penalty.

PPO compares the current probability of the recorded pair with the probability under the policy that collected it.
The ratio $\rho_t$ in \Cref{eq:joint-ratio} exceeds one when that pair has become more likely.
The denominator is fixed throughout the update.
For example, changing the old joint probability $.336$ to a current probability $.420$ gives $\rho=1.25$.
With positive advantage $.5$ and clipping parameter $\eta=.2$, the two candidate contributions are $1.25(.5)=.625$ and $1.2(.5)=.6$.
PPO uses their minimum, $.6$, so the minimized actor loss receives $-.6$ from this sample.

Clipping limits the surrogate's incentive to keep increasing an already-favored sampled decision.
For negative advantages it similarly limits the incentive to keep decreasing a disfavored decision beyond the lower ratio threshold.
Shared parameters and the other loss terms can still change the ratios after an update; clipping acts on the actor surrogate.
The clipped objective is written explicitly in \Cref{eq:ppo-explicit}.

\subsection{The PPO surrogate and shared target convention}
\label{app:training-math}
A rollout records both the external action and the internal depth that produced it. Keeping the depth makes it possible to assign likelihood-ratio credit to stopping as well as acting.
Freeze a rollout policy $\theta_{\rm old}$ and store $(s_t,\tau_t,a_t,r_t,s_{t+1})$ together with its joint log-probability. With fixed advantages $\widehat A_t$, the actor loss is
\begin{equation}
\cL_{\rm PPO}(\theta)=-\widehat\E_t\!\left[
\min\!\left\{\rho_t(\theta)\widehat A_t,
\operatorname{clip}(\rho_t(\theta),1-\eta,1+\eta)\widehat A_t\right\}\right],
\label{eq:ppo-explicit}
\end{equation}
where $\eta>0$ is the clipping parameter (default $0.2$) and
\begin{equation}
\rho_t(\theta)=
\frac{q_\theta(\tau_t\mid s_t)\pi_{\theta,\tau_t}(a_t\mid s_t)}
{q_{\theta_{\rm old}}(\tau_t\mid s_t)\pi_{\theta_{\rm old},\tau_t}(a_t\mid s_t)}.
\label{eq:joint-ratio}
\end{equation}
The denominator is the frozen rollout policy's joint probability, or density for continuous actions and is nonzero on stored samples.
Removing its stopping factor would remove direct likelihood-ratio credit for the sampled stopping decision.
Although the environment observes only the marginal action policy $\pi_\theta(a\mid s)=\sum_dq_\theta(d\mid s)\pi_{\theta,d}(a\mid s)$, the recorded depth supports the joint estimator.

The shared GAE target is defined using the frozen, state-only bootstrap baseline $b_{\rm old}(s)$ used by the rollout implementation. Let $m_t=0$ after a true terminal transition and $m_t=1$ otherwise and let $\lambda_{\rm GAE}\in[0,1]$ (default $0.95$) be distinct from the loss coefficients. For a rollout segment ending at index $T-1$, the convention is
\begin{align}
\delta_t&=r_t+\gamma m_t b_{\rm old}(s_{t+1})-b_{\rm old}(s_t),\\
\widehat A_t&=\sum_{\ell=0}^{T-t-1}(\gamma\lambda_{\rm GAE})^\ell
\left(\prod_{k=0}^{\ell-1}m_{t+k}\right)\delta_{t+\ell},
\qquad V_t^{\rm targ}=b_{\rm old}(s_t)+\widehat A_t.
\label{eq:gae-shared}
\end{align}
A nonterminal rollout cutoff bootstraps from $b_{\rm old}(s_T)$.
Segments end at environment resets. If a time limit truncates a continuing task, bootstrap from its final pre-reset observation, but do not extend the advantage trace into the reset episode.
A baseline that depends on the sampled $\tau_t$ is not automatically valid for the joint score estimator.
Targets, old probabilities and advantages remain fixed during a PPO update and every depth critic receives the same $V_t^{\rm targ}$ in \Cref{eq:full-loss}.

\paragraph{Local cost surrogate versus trajectory cost.}
The expectation of $C_\theta(s_t)$ in \Cref{eq:full-loss} uses frozen rollout states. It differentiates the conditional depth distribution at those states, with no derivative through their sampling distribution. In contrast, the full discounted objective
\begin{equation}
J_c(\theta)=\E_{p_\theta}\sum_{t\ge0}\gamma^t
\left(r_t-\lambda_c\frac{\tau_t}{\dmax}\right)
\label{eq:trajectory-compute-objective}
\end{equation}
also changes future state visits and future compute. A policy-gradient treatment of this objective would use return/advantage information including those future computation costs. Reward-only GAE plus a conditional cost penalty does not in general give that exact gradient. The same units yield the statewise coefficient $c=\lambda_c/\dmax$, but PPO clipping, advantage scaling, value training and distribution shift prevent identifying the training loss with the ideal stopping optimum.

\subsection{Loss terms}
\label{app:loss-walkthrough}
The full loss in \Cref{eq:full-loss} combines an actor term, an auxiliary value term and a computation term.
The actor term connects stopping and action selection to task outcomes.
The value term supplies a prediction task at every representation.
The computation term expresses a preference for spending fewer refinements when the reward objective does not justify them.

\paragraph{Value supervision.}
At each recorded state, the squared error $(\widehat v_{\theta,d}-V_t^{\rm targ})^2$ compares a depth's prediction with the common target.
Averaging over depths keeps the total value-loss weight from scaling automatically with the number of heads.
Even if depth two supplied the executed action, the critics at depths one, three and later depths still receive this loss.
Their gradients also pass through the shared representations and refinement block.
This can change action quality indirectly, which is why the factorial experiment separates auxiliary supervision from adaptive allocation.
Supervising every value head does not imply that every action head receives a direct actor update on that sample.

\paragraph{Computation penalty.}
The normalized expectation $C_\theta(s)$ lies between $1/\dmax$ and one, because at least one refinement is performed and the cap is forced.
It is differentiable through the current stopping probabilities.
For the four-depth calculation above, the penalty is $\lambda_c(.59)$; if $\lambda_c=.01$, it contributes $.0059$ before averaging across states.
The expectation includes all possible stopping depths at that state, even though only one depth was sampled for the trajectory.
Increasing an early stopping probability reduces the chance of paying for later refinements.
\Cref{lem:stopping-law} gives the exact derivative and covers probabilities at the boundary.

A sampled integer depth is a recorded outcome, not a differentiable probability.
Treating that integer as a constant and differentiating its numerical value supplies no gradient to the halting network.
The analytic expectation avoids this problem.
A separate score-function estimator is possible, as shown in \Cref{eq:cost-score}, but its sampling distribution and any importance weights must be handled consistently.

\paragraph{Value differences.}
The difference $\widehat v_d-\widehat v_{d-1}$ compares two predictions of the same target.
It is not a target for the halting head and need not equal the return gain from forcing another refinement.
\Cref{prop:shared-target} derives the prediction target and gives a one-state example.
The halting head receives task credit from the joint actor loss and depth credit from the computation penalty.

\subsection{Shared-target regression}
\label{app:shared-target-proof}
The auxiliary loss teaches each representation to predict the rollout target. It does not replace that target with the outcome of forcing a particular depth. Here $W_d(s)$ denotes the expected return after choosing an action from $\pi_d$ at state $s$ and then following a fixed continuation policy $\bar\pi$.
Its action value $Q^{\bar\pi}(s,a)$ includes the immediate reward and subsequent discounted rewards.
The distinction follows directly from the population minimizer of squared error.

\begin{proposition}[Shared targets do not identify forced-depth values]
\label{prop:shared-target}
Fix a joint distribution of a state $S$ and a square-integrable shared target $Y$ and fix each representation $Z_d=z_d(S)$. The minimizer of $\E[(f_d(Z_d)-Y)^2]$ over square-integrable measurable predictions is $f_d^*(Z_d)=\E[Y\mid Z_d]$, almost surely. This conditional mean need not equal $W_d(S)=\E_{a\sim\pi_d}[Q^{\bar\pi}(S,a)]$.
\end{proposition}
\begin{proof}
\emph{Step 1: separate predictable signal from residual error.}
Fix a depth and write $Z=Z_d$, $m(Z)=\E[Y\mid Z]$ and $f(Z)=f_d(Z_d)$.
The conditional mean $m(Z)$ is the average target among samples with the same representation.
Write the prediction error as
\[
f(Z)-Y=\underbrace{f(Z)-m(Z)}_{\text{prediction error relative to the mean}}
+\underbrace{m(Z)-Y}_{\text{residual around that mean}}.
\]
By the definition of a conditional mean, $\E[m(Z)-Y\mid Z]=0$.
Because $f(Z)-m(Z)$ is determined by $Z$, the cross term in the squared error satisfies
\[
\E[(f(Z)-m(Z))(m(Z)-Y)]
=\E[(f(Z)-m(Z))\E[m(Z)-Y\mid Z]]=0.
\]

\emph{Step 2: identify the minimizer.}
Expand the square and use the zero cross term:
\begin{equation}
\E[(f_d(Z_d)-Y)^2]
=\E[(f_d(Z_d)-m_d(Z_d))^2]
 +\E[(Y-m_d(Z_d))^2].
\label{eq:regression-projection}
\end{equation}
The second term is fixed by the data distribution and representation.
The first term is nonnegative and becomes zero exactly when $f_d(Z_d)=m_d(Z_d)$ almost surely.
This proves the population minimizer.
It concerns the target $Y$ actually supplied to regression; the calculation has not changed that target into an outcome obtained by forcing depth $d$.

\emph{Step 3: show why the two value notions can disagree.}
Consider a one-state bandit that ends after one action.
At depth one, $\pi_1$ always selects an action with reward zero.
At depth two, $\pi_2$ always selects an action with reward one.
During data collection, choose depth two with probability $p\in(0,1)$ and give both critics the observed reward $Y$.
Then
\[
\Pr(Y=1)=p,\qquad \Pr(Y=0)=1-p,\qquad \E[Y]=p.
\]
Both representations are constant, so conditioning on either representation leaves that mean unchanged.
The optimal predictions are therefore $f_1^*=f_2^*=p$.
By contrast, executing each action head gives $W_1=0$ and $W_2=1$.
Hence
\[
f_2^*-f_1^*=0\quad\text{while}\quad W_2-W_1=1.
\]
Even exact shared-target regression can therefore give value differences that do not equal forced-depth return differences.
\end{proof}

When every representation retains all state information, the optimal predictions equal $\E[Y\mid S]$ at every depth.
Information loss can make these conditional means differ; finite model classes, shared parameters, bootstrapping and optimization introduce further differences.
Estimating forced-depth returns requires targets for those interventions.
The joint policy ratio supplies reward credit to the sampled action and stopping path.

\subsection{Invariance to value outputs}
\label{app:calibration}
The architecture separates a training pathway from an inference pathway.
During training, the loss on $V_d(z_d)$ updates the critic and shared representation.
The updated representation can change both $\pi_d(z_d)$ and $h_d(z_d)$.
At inference, the frozen policy computes actions and stopping directly from the representation.
It does not read the numerical critic outputs.

\begin{proposition}[Invariance to value-output replacement]
\label{prop:value-output-invariance}
Fix the encoder, refinement block, action heads and halting heads.
Replace each auxiliary value output by an arbitrary function $\widetilde V_d(z_d)$ without modifying any of those fixed mappings or the stopping algorithm.
Then the joint action/depth law $p_\theta(a,d\mid s)$ in \Cref{eq:joint-policy}, the deterministic threshold stopping depth and the expected environment return are unchanged.
\end{proposition}
\begin{proof}
\emph{Step 1: trace one forward decision.}
The input and encoder are fixed, so $z_0$ is unchanged.
Inductively, an unchanged $z_{d-1}$ and refinement block produce the same $z_d$.
The fixed action and halting heads then produce the same $\pi_d$ and $h_d$ at every depth.
Replacing only $V_d(z_d)$ affects none of these mappings.

\emph{Step 2: compare the two stopping conventions.}
Under stochastic stopping, the products defining $S_d$ and $q(d\mid s)$ use only the unchanged hazards.
Thus both $q(d\mid s)$ and $q(d\mid s)\pi_d(a\mid s)$ are unchanged.
Under thresholded evaluation, the first depth with $h_d>1/2$, or the forced cap, is also unchanged.
Its selected action distribution is the same.

\emph{Step 3: extend the equality to trajectories.}
Begin with the same initial-state distribution and environment dynamics.
At the first state, identical action kernels give the same distribution of the next state and reward.
Repeating this argument at each step gives identical finite trajectory distributions.
Their expected returns are therefore equal whenever those returns are defined, including bounded-horizon and bounded-reward discounted settings.
\end{proof}

This result concerns a frozen forward policy.
Changing value targets during training can change the shared representation and is therefore a different intervention.
Replacing the stopping rule by a threshold on value increments would also define a different evaluation policy.

\subsection{Training loop}
\label{app:update-walkthrough}
A rollout is a sequence of environment interactions collected before updating the policy.
For one update cycle, the parameters used to collect that data are denoted by $\theta_{\rm old}$ and kept fixed as the reference policy.
For each state, the training procedure computes the full refinement chain, samples a depth from $q_{\theta_{\rm old}}$ and samples the action from the selected head.
Only that action is executed.
The other representations remain available for auxiliary value supervision; computing them does not create additional environment transitions.

The stored record contains the state, selected depth, executed action, reward, next state, termination information and old joint log-probability.
The rollout baseline and subsequent rewards then determine the fixed advantages and value targets.
Old joint probabilities and targets must remain associated with the same recorded transition when samples are batched or shuffled.
Otherwise the likelihood ratio or regression target would describe a different decision.

For a batch of $N$ recorded states, the stopping distribution can be viewed as an array with $N$ rows and $\dmax$ columns: each row is a normalized distribution over depths.
The value predictions form another array of the same shape, but all entries in a row are compared with one shared target for that state.
The actor likelihood selects the recorded depth's action head for each row.
The depth penalty sums $d\,q_\theta(d\mid s)/\dmax$ along that row.

Each optimization pass recomputes the current representations, probabilities and value predictions on the recorded states.
The three losses are averaged and combined as in \Cref{eq:full-loss} and differentiation updates the trainable parameters.
The old-policy quantities remain fixed while the current quantities change.
After the update cycle, new trajectories are collected with the updated policy.
\Cref{alg:aic} summarizes these operations; \Cref{app:training-math} distinguishes this rollout-state cost surrogate from a full trajectory objective that includes future computation costs.

\begin{algorithm}[h]
\caption{AIC: full-depth training and adaptive-depth evaluation.}
\label{alg:aic}
\footnotesize
\textbf{Inputs:} encoder, shared refinement block, depth-specific heads, $\dmax$, $\lambda_c$, $\lambda_V$.
\begin{enumerate}[label=\arabic*.,leftmargin=1.8em,itemsep=2pt,topsep=3pt]
\item Freeze rollout parameters $\theta_{\rm old}$. For each state, compute $z_0$ and all $z_d,\pi_d,V_d$; compute $h_d$ for $d<\dmax$.
\item Sample $\tau\sim q_{\theta_{\rm old}}(\cdot\mid s)$, with a forced stop at $\dmax$ and sample $a\sim\pi_{\theta_{\rm old},\tau}$.
\item Execute $a$; store the transition, $\tau$ and $\log q_{\theta_{\rm old}}(\tau\mid s)+\log\pi_{\theta_{\rm old},\tau}(a\mid s)$.
\item Construct frozen rollout advantages and a shared value target at every state as in \Cref{eq:gae-shared}.
\item For each PPO update, recompute all depths and the joint ratio in \Cref{eq:joint-ratio}. Minimize \Cref{eq:full-loss}, differentiating the expected normalized depth through $q_\theta$.
\item At evaluation, refine sequentially and stop at the first $d<\dmax$ with $h_d>0.5$, or at the cap; execute an action from $\pi_{\theta,d}$.
\end{enumerate}
\end{algorithm}

\subsection{Deployment and computation accounting}
\label{app:deployment-walkthrough}
Deployment uses the learned network to process one state sequentially.
It encodes the input, applies the shared refinement block and stops at the first depth below the cap for which $h_d>0.5$.
If no such depth occurs, it stops at $\dmax$.
The selected action head then supplies the action.
Later refinements need not be evaluated to select that action, because all earlier stopping decisions depend only on representations already computed.

In the four-depth example, $h_1=.2$ continues and $h_2=.6$ stops, giving evaluation depth two compared with stochastic mean depth $2.36$.
The action is drawn from the selected action distribution.

\paragraph{Thresholding is a different evaluation policy.}
Define $\tau_{\rm eval}(s)=\min(\{d<\dmax:h_d(s)>1/2\}\cup\{\dmax\})$. This is neither the mean nor necessarily the mode of $q_\theta$. For example, $\dmax=3$ and $h_1=h_2=0.4$ give $q=(0.4,0.24,0.36)$, with mean depth $1.96$, modal depth one and thresholded depth three. Near-binary hazards can reduce the discrepancy. We measure inference computation and return under the deployed threshold policy.

The expected-depth penalty describes refinement usage, not all execution costs.
Encoding, output heads, memory movement and control flow also take time.
An expensive frozen encoder can dominate runtime even when many refinements are skipped and batching can affect whether different stopping depths translate into latency savings.
Training computes all depths for supervision, so an early sampled stopping depth does not imply the same saving during training.
These distinctions motivate the separate measurements of depth, inference latency and training time in \Cref{app:efficiency}.

Finally, the allocation theory studies what could be gained if depth were assigned using known depth-specific action values under its assumptions.
Those values, denoted by $W_d$, are expected returns from selecting a particular depth's action and then following a fixed continuation policy.
They differ from the shared-target predictions $\widehat v_d$.
AIC learns a stopping policy from sampled outcomes and the experiments evaluate the resulting reward and computation.

\section{Theory of State-Dependent Computation Allocation}
\label{app:proofs}
We study whether assigning computation to the states that need it can improve value without increasing average depth.
We answer this question for a fixed state distribution and a fixed family of depth-specific action policies.
The comparison changes only how depth is assigned.

\subsection{Setup, units and assumptions}
\label{app:proofs-setup}
Let $\mu$ be a fixed probability distribution over states and let the integer depth cap satisfy $\dmax\ge K\ge2$. A stopping law $q(d\mid s)$ is a probability distribution on $\{1,\ldots,\dmax\}$. For integrable decision values $W_d(s)$ and an integrable reference $W^*(s)\ge W_d(s)$, define
\begin{align}
\mathcal B_\mu(q)&:=\E_{s\sim\mu}\sum_{d=1}^{\dmax}d\,q(d\mid s),\label{eq:fixed-mu-budget}\\
\mathcal R_\mu(q)&:=\E_{s\sim\mu}\sum_{d=1}^{\dmax}q(d\mid s)\big[W^*(s)-W_d(s)\big].\label{eq:fixed-mu-regret}
\end{align}
The budget $\mathcal B_\mu$ counts expected refinement steps under $\mu$. The training penalty instead averages the statewise normalized cost $C_\theta(s)=\E_\theta[\tau\mid s]/\dmax$ over rollout states.
The mean-depth budget $b$ used below is therefore distinct from the nominal experimental cap $B$.
The expectation over $s\sim\mu$ averages over states, while the sum over $d$ averages over the stopping choice at a given state.
Thus $\mathcal B_\mu$ is average work and $\mathcal R_\mu$ is average value lost relative to the reference.
All allocations use the same fixed action-policy family and continuation policy, with no retraining.
Here $W_d$ is a decision value used for analysis, not an auxiliary prediction $\widehat v_d$.
The following assumptions describe when easy states can stop early while hard states need additional depth.

\paragraph{(A1) A measurable difficulty partition.}
The state space is partitioned into easy states $E$ and hard states $H$, with $\mu(H)=\epsilon\in(0,1)$ and $\mu(E)=1-\epsilon$.
\paragraph{(A2) Attainability.}
For $s\in E$, $W_d(s)=W^*(s)$ at every depth. For $s\in H$, $W_d(s)=W^*(s)$ at every $d\ge K$.
\paragraph{(A3) Shortfall before sufficient depth.}
There is a scale $\Delta>0$ such that for every $s\in H$ and $1\le d<K$,
\begin{equation}
W^*(s)-W_d(s)\ge\Delta(1-d/K).
\label{eq:depth-shortfall}
\end{equation}
All statements hold $\mu$-almost surely.
The scale $\Delta$ controls shortfall before sufficient depth; it is not a separately assumed action gap.
Assumptions (A2)--(A3) make $K$ the first optimal depth on a hard state and the ideal allocator knows the easy/hard partition.
The allocation result is conditional on these decision-value assumptions.

\subsection{A tight worked example}
The benefit comes from associating depth with the state.
The following construction makes this association concrete and will attain the theorem's bound exactly.
Consider four equally likely states, which we call contexts in this one-decision example. Three have $W_d=1$ at every depth; the fourth has $W_d=d/4$ for $d=1,\ldots,4$. Set $K=\dmax=4$, $\epsilon=1/4$ and $\Delta=1$. The ideal allocator uses depths $(1,1,1,4)$, so its mean value is one and its mean depth is $b=7/4$.

The adaptive rule spends one step on each of the three easy contexts and four on the hard context.
A state-independent rule cannot direct its deeper choices toward the hard context: it uses the same depth distribution in all four contexts.

A feasible deterministic constant-depth baseline uses depth one, with mean value $(3+1/4)/4=13/16=0.8125$. Every state-independent schedule with mean depth $7/4$ has mean value
\begin{equation}
\frac34+\frac14\,\E[\tau/4]
=\frac34+\frac{7}{64}
=\frac{55}{64}=0.859375.
\end{equation}
For example, choose depths one and two with probabilities $1/4$ and $3/4$. Its shortfall is $9/64=0.140625$, exactly the bound $\epsilon\Delta(1-b/K)$.
This finite-context construction attains the lower bound in \Cref{eq:allocation-lower-bound}.

\subsection{\texorpdfstring{Proof of \Cref{thm:main}, including randomized schedules}{Proof of Theorem 1, including randomized schedules}}
The proof has two parts: construct an allocation with zero shortfall, then show that every state-independent schedule at its mean-depth budget has positive shortfall.
Randomization is allowed in the baseline, so the argument does not rely on comparing a fractional mean depth with an integer fixed depth.
\begin{proof}
\emph{Step 1: construct the state-dependent allocation.}
Let $q^*$ stop at depth one on the easy set $E$ and at depth $K$ on the hard set $H$.
Assumption (A2) makes both choices attain $W^*(s)$, so every statewise shortfall is zero and $\mathcal R_\mu(q^*)=0$.
Its expected work is
\[
\mathcal B_\mu(q^*)=(1-\epsilon)\cdot1+\epsilon K
=b:=1+\epsilon(K-1).
\]
The easy states make this budget smaller than $K$:
$K-b=(1-\epsilon)(K-1)>0$.

\emph{Step 2: lower-bound shortfall for any chosen depth.}
Write $(x)_+=\max\{x,0\}$.
For a hard state and $d<K$, (A3) gives shortfall at least $\Delta(1-d/K)$.
For $d\ge K$, (A2) gives zero shortfall and $(1-d/K)_+=0$.
Both cases are therefore captured by
\[
W^*(s)-W_d(s)\ge\Delta(1-d/K)_+\qquad(s\in H).
\]
Easy states have zero shortfall.

\emph{Step 3: average under a state-independent schedule.}
Let $\tau\sim\nu$ have $\E_\nu[\tau]\le b$ and use this same distribution at every state.
The depth is independent of whether the sampled state is hard.
Since hard states have probability $\epsilon$, Step 2 gives
\[
\mathcal R_\mu(\nu)
\ge\E_{s\sim\mu,\,\tau\sim\nu}
   [\mathbf 1\{s\in H\}\Delta(1-\tau/K)_+]
=\epsilon\Delta\,\E_{\tau\sim\nu}[(1-\tau/K)_+].
\]
Independence is essential to this factorization; it need not hold for a state-dependent rule.

\emph{Step 4: use only the baseline's average budget.}
The function $f(x)=(1-x/K)_+$ is the maximum of two affine functions, $1-x/K$ and zero and is therefore convex.
Jensen's inequality gives
\begin{align}
\mathcal R_\mu(\nu)
&\ge\epsilon\Delta\,\E[f(\tau)]
\ge\epsilon\Delta\,f(\E[\tau])\nonumber\\
&\ge\epsilon\Delta\,f(b)
=\epsilon\Delta(1-b/K)>0.
\label{eq:randomized-lower-bound}
\end{align}
The third inequality uses that $f$ is non-increasing and $\E[\tau]\le b$.
The last step uses $b<K$ from Step 1.
This proof also allows baseline mass above $K$: the positive-part function already includes those depths.
\end{proof}

\paragraph{Deterministic fixed-depth special case.}
For a constant integer depth $K'\le b$, feasibility requires $K'\le\lfloor b\rfloor<K$. Directly applying (A3) gives the sharper depth-specific bound
\begin{equation}
\mathcal R_\mu(K')\ge\epsilon\Delta(1-K'/K).
\label{eq:constant-depth-bound}
\end{equation}
The comparison uses an upper compute budget; an integer baseline need not consume exactly $b$. A distribution over $\lfloor b\rfloor$ and $\lceil b\rceil$ can consume $b$ exactly and \Cref{eq:randomized-lower-bound} still applies.

\subsection{More than two difficulty levels and the saturated case}
The same argument can be applied separately to groups that need different depths.
The resulting lower bound adds the contribution from each group.

\begin{proposition}[Multiple difficulty levels]
\label{prop:conc}
Suppose the hard set is partitioned into finitely many levels $H_i$ with conditional masses $p_i>0$, $\sum_i p_i=1$, earliest optimal depths $2\le K_i\le\dmax$ and shortfall scales $\Delta_i>0$. Replace (A2)--(A3) on $H_i$ by their versions with $K_i,\Delta_i$. The ideal state-dependent allocator has zero shortfall and mean depth
\begin{equation}
b^*=1+\epsilon\sum_i p_i(K_i-1).
\end{equation}
Every state-independent schedule with $\E[\tau]\le b^*$ obeys
\begin{equation}
\mathcal R_\mu\ge\epsilon\sum_i p_i\Delta_i(1-b^*/K_i)_+.
\label{eq:multi-level-bound}
\end{equation}
A constant depth $K'$ instead obeys the bound with $b^*$ replaced by $K'$.
\end{proposition}
\begin{proof}
\emph{Step 1: construct the allocation and compute its budget.}
Use depth one on $E$ and depth $K_i$ on $H_i$.
Each group then attains its reference value, giving zero shortfall.
Because $\mu(H_i)=\epsilon p_i$, the mean depth is
\[
(1-\epsilon)\cdot1+\sum_i\epsilon p_iK_i
=1+\epsilon\sum_i p_i(K_i-1)=b^*.
\]

\emph{Step 2: bound each group's contribution.}
For a state-independent depth $\tau$, the shortfall on $H_i$ is at least $\Delta_i(1-\tau/K_i)_+$.
Independence permits averaging over depths before weighting by the group's probability:
\[
\mathcal R_\mu
\ge\epsilon\sum_i p_i\Delta_i\E[(1-\tau/K_i)_+].
\]
For each $i$, convexity and the budget condition imply
\[
\E[(1-\tau/K_i)_+]
\ge(1-\E[\tau]/K_i)_+
\ge(1-b^*/K_i)_+.
\]
Substituting this inequality into the sum proves \Cref{eq:multi-level-bound}.
For a constant depth $K'$, substitute $\tau=K'$ directly instead.
\end{proof}

The bound can be zero for a group with $K_i\le b^*$; no positive separation is claimed from that group alone.
At a fixed baseline depth, its summand increases with $K_i$.
Changing $K_i$ also changes the ideal budget $b^*$, so this observation does not imply unrestricted monotonicity when both are varied.

At the other extreme, if all depths produce the same decision value, there is no reward benefit to offset extra work.

\begin{corollary}[No benefit from depth]
\label{cor:saturated}
If $W_d(s)=W_1(s)$ at every state and depth, a strictly positive cost $c$ per refinement step makes depth one the unique optimal stopping depth in the ideal statewise problem.
\end{corollary}
\begin{proof}
For every $d>1$, $[W_d(s)-cd]-[W_1(s)-c]=-c(d-1)<0$. A distribution placing positive mass on such a depth has lower utility than stopping at one. This gives a unique stopping distribution, although hazards after the certain stop are arbitrary.
\end{proof}
DistShift's saturated result is compatible with this condition; the statement does not guarantee that training finds this optimum.

\subsection{Sequential returns}
\label{app:occupancy}
The fixed-$\mu$ analysis compares decisions on a common state distribution.
In a sequential MDP, changing the stopping policy can also change which states are visited.
The next identity makes the required state distribution and continuation values explicit before relating shortfall to total return.

\begin{lemma}[Discounted occupancy identity]
\label{lem:occupancy}
Consider a discounted MDP with bounded rewards, discount $0\le\gamma<1$ and a fixed initial-state distribution. Let $\pi^*$ be an optimal policy and let $\pi_q(a\mid s)=\sum_d q(d\mid s)\pi_d(a\mid s)$. Define the normalized discounted occupancy
\[
d_{\pi_q}(s)=(1-\gamma)\sum_{t\ge0}\gamma^t\Pr_{\pi_q}(s_t=s),
\]
with the analogous measure definition for non-discrete states. If $W^*=V^*$ and $W_d(s)=\E_{a\sim\pi_d}[Q^*(s,a)]$, then
\begin{equation}
J(\pi^*)-J(\pi_q)=\frac{\mathcal R_{d_{\pi_q}}(q)}{1-\gamma}.
\label{eq:occupancy-gap}
\end{equation}
\end{lemma}
\begin{proof}
\emph{Step 1: express shortfall as a one-transition residual.}
Define the statewise shortfall
$\ell_q(s)=V^*(s)-\sum_dq(d\mid s)W_d(s)$.
The Bellman definition of $Q^*$ gives
\[
\E[r_t+\gamma V^*(s_{t+1})\mid s_t=s]
=\E_{a\sim\pi_q}[Q^*(s,a)]
=\sum_dq(d\mid s)W_d(s).
\]
Therefore
$\ell_q(s_t)=\E[V^*(s_t)-r_t-\gamma V^*(s_{t+1})\mid s_t]$.

\emph{Step 2: add the residuals along a trajectory.}
Multiply by $\gamma^t$, sum through $T$ and take expectation under $\pi_q$:
\begin{align*}
\E_{\pi_q}\sum_{t=0}^{T}\gamma^t\ell_q(s_t)
&=\E_{\pi_q}\sum_{t=0}^{T}
 [\gamma^tV^*(s_t)-\gamma^tr_t-\gamma^{t+1}V^*(s_{t+1})]\\
&=\E[V^*(s_0)]
-\E_{\pi_q}\sum_{t=0}^{T}\gamma^tr_t
-\gamma^{T+1}\E_{\pi_q}[V^*(s_{T+1})].
\end{align*}
The value terms at $s_1,\ldots,s_T$ cancel in pairs.
If rewards are bounded in magnitude by $R_{\max}$, then
$|V^*(s)|\le R_{\max}/(1-\gamma)$.
The remaining terminal term is bounded by
$\gamma^{T+1}R_{\max}/(1-\gamma)$ and tends to zero.
The limit is $J(\pi^*)-J(\pi_q)$, since the initial-state distribution is the same for both policies.

\emph{Step 3: rewrite the trajectory sum as a state average.}
The factor $(1-\gamma)$ normalizes the discounted occupancy because
$(1-\gamma)\sum_{t\ge0}\gamma^t=1$.
By its definition,
\[
\E_{s\sim d_{\pi_q}}[\ell_q(s)]
=(1-\gamma)\E_{\pi_q}\sum_{t\ge0}\gamma^t\ell_q(s_t)
=\mathcal R_{d_{\pi_q}}(q).
\]
Combining this equality with Step 2 and dividing by $1-\gamma$ proves \Cref{eq:occupancy-gap}.
\end{proof}

The same occupancy enters the deployed compute budget:
$(1-\gamma)\E_{\pi_q}\sum_{t\ge0}\gamma^t\tau_t=\mathcal B_{d_{\pi_q}}(q)$.
Because two policies can have different occupancies, equal computation under an arbitrary $\mu$ need not mean equal computation along their trajectories.
Applying the theorem to sequential returns therefore requires the value-profile assumptions and hard-state mass under the relevant occupancies, together with a compute check for each policy.
The identity also uses $Q^*$: a general common continuation value $Q^{\bar\pi}$ cannot replace it without further assumptions.

\section{Optimal Stopping for a Fixed Value Profile}
\label{app:halt-derivation}
If the return available at each depth were known at one state, when should computation stop?
The analysis concerns fixed decision values; AIC learns a halting policy from sampled outcomes rather than receiving this profile as input.

\subsection{A fixed-policy, fixed-state subproblem}
Fix a state $s$, all depth-specific action distributions $\pi_d$ and a common continuation policy $\bar\pi$ at subsequent environment steps. Define
\begin{equation}
W_d(s)=\E_{a\sim\pi_d(\cdot\mid s)}[Q^{\bar\pi}(s,a)],\qquad
U_d(s;c)=W_d(s)-cd,\quad c\ge0.
\label{eq:halt-utility}
\end{equation}
The quantity $W_d$ changes the current action distribution while holding future behavior fixed. No environment transition or additional temporal discount occurs during a refinement step. The profile is treated as fixed and available to the ideal decision maker. It is distinct from the learned shared-target predictions in \Cref{eq:voc}.
At the same unscaled reward units, the normalized depth penalty corresponds to $c=\lambda_c/\dmax$.

\subsection{Constant cost and adaptive depth}
\label{app:voc-vs-ponder}
State-dependent stopping does not require a state-dependent price for computation.
With fixed values $W_d(s)$ and a constant price $c>0$, the ideal decision is $\max_d[W_d(s)-cd]$.
Its maximizing depth can vary because the benefit of refinement varies across states.
This also applies to ACT-style ponder objectives: a flat cost is compatible with adaptive computation.

In the four-context construction of \Cref{app:proofs}, every additional step on an easy context has value increment zero, whereas every additional step on the hard context has value increment $1/4$. For $0<c<1/4$, utility strictly decreases with depth on the easy contexts and strictly increases on the hard one. The unique optimal depths are therefore one and four, respectively, giving precisely the ideal allocation. At $c=1/4$ all hard-context depths tie; for $c>1/4$ depth one is optimal everywhere. Thus the strict interval on $c$ is necessary for that example's unique adaptive solution.

The learning problem is to infer the useful depth from information available before stopping.
AIC addresses that problem through joint RL training and tests when the resulting allocation explains performance gains.

\subsection{The general optimal continuation condition}
At a given depth, continuing is useful when the best later outcome compensates for the remaining computation cost. The next recursion compares stopping now with that full continuation opportunity.

\begin{lemma}[Finite-depth stopping recursion]
\label{lem:bellman-stop}
After reaching depth $d$, treat its computation as sunk and define
\begin{equation}
F_d(s)=\max_{d\le k\le\dmax}\{W_k(s)-c(k-d)\}.
\label{eq:remaining-value}
\end{equation}
Then $F_{\dmax}=W_{\dmax}$ and, for $d<\dmax$,
\begin{equation}
F_d(s)=\max\{W_d(s),\,F_{d+1}(s)-c\}.
\label{eq:stop-bellman}
\end{equation}
Stopping at $d$ is optimal among remaining depths exactly when $W_d(s)\ge F_{d+1}(s)-c$; equality permits either stopping or an optimal continuation.
\end{lemma}
\begin{proof}
\emph{Step 1: handle the final depth.}
At $d=\dmax$, there is no remaining refinement.
The only feasible choice in \Cref{eq:remaining-value} is $k=\dmax$, so $F_{\dmax}=W_{\dmax}$.

\emph{Step 2: split the remaining choices into stop and continue.}
At $d<\dmax$, choosing $k=d$ yields $W_d$ with no additional computation cost.
For every later choice $k\ge d+1$, separate the cost of the next step:
\[
W_k-c(k-d)
=\big[W_k-c(k-(d+1))\big]-c.
\]
The maximum of the bracket over $k\ge d+1$ is exactly $F_{d+1}$.
Taking the larger of the stop and continue options gives
$F_d=\max\{W_d,F_{d+1}-c\}$.
Stopping is therefore optimal exactly when $W_d\ge F_{d+1}-c$.
Equality means that both options attain the same value.

\emph{Step 3: reconcile remaining cost with total cost.}
At depth $d$, the first $d$ steps have already been paid for.
Subtracting their constant cost $cd$ gives
\[
F_d-cd=\max_{k\ge d}\{W_k-c(k-d)-cd\}
=\max_{k\ge d}(W_k-ck).
\]
In particular, $\max_kU_k=F_1-c$.
The first refinement costs $c$ too; it does not affect which depth maximizes utility.
\end{proof}
Without additional structure, the relevant benefit is the best continuation value, not just $W_{d+1}-W_d$.
For stochastic internal transitions, continuation would require a conditional expectation over future internal states.

\subsection{First-crossing rule}
A one-step threshold is sufficient when successive benefits diminish. Once an increment falls below cost, no later increment can make continued refinement worthwhile under this assumption.

\begin{proposition}[Diminishing returns is a sufficient condition]
\label{prop:halt-optimal}
Let $g_d(s)=W_{d+1}(s)-W_d(s)$ for $d<\dmax$. If $g_d(s)$ is non-increasing in $d$, the earliest maximizing depth is
\begin{equation}
\dstar(s)=\min\big(\{d<\dmax:g_d(s)\le c\}\cup\{\dmax\}\big).
\label{eq:dstar-rule}
\end{equation}
\end{proposition}
\begin{proof}
\emph{Step 1: translate value increments into utility increments.}
Since $U_d=W_d-cd$,
\[
U_{d+1}-U_d=(W_{d+1}-W_d)-c=g_d-c.
\]
The assumption that $g_d$ is non-increasing also makes these utility increments non-increasing.

\emph{Step 2: compare the first crossing with every earlier depth.}
Suppose the threshold set is nonempty and let $d_0$ be its first element.
Before $d_0$, each $g_j-c$ is strictly positive.
For any $k<d_0$, adding those increments gives
\[
U_{d_0}-U_k=\sum_{j=k}^{d_0-1}(g_j-c)>0.
\]
Thus no earlier depth is optimal.

\emph{Step 3: compare it with every later depth.}
At the crossing, $g_{d_0}\le c$.
Non-increasing increments imply $g_j\le c$ for every $j\ge d_0$.
Hence for any $k>d_0$,
\[
U_k-U_{d_0}=\sum_{j=d_0}^{k-1}(g_j-c)\le0.
\]
No later depth has greater utility, although zero increments can create ties.
This proves that $d_0$ is the earliest global maximizer.

\emph{Step 4: cover the cap and single-depth cases.}
If the threshold set is empty, every utility increment is positive and the cap is the unique maximizer.
If $\dmax=1$, there are no increments and depth one is the only feasible choice.
The stated formula covers both cases.
\end{proof}
With $\Delta W_{d+1}:=W_{d+1}-W_d$, the familiar condition
\begin{equation}
\Delta W_{d+1}(s)\le c
\label{eq:voc-threshold}
\end{equation}
is therefore sufficient under diminishing returns. It is not a general stopping condition for arbitrary profiles and evaluating its next-depth counterfactual can itself require computation.

\paragraph{Counterexample without diminishing returns.}
Let $(W_1,W_2,W_3)=(0,0,3)$ and $c=1$. The first increment is zero, so a first-crossing heuristic stops at one. The utilities are $(U_1,U_2,U_3)=(-1,-2,0)$ and depth three is uniquely optimal. The recursion continues because the later benefit compensates for both extra steps. This is why an observed low immediate increment does not by itself justify a learned early exit.

\subsection{Optimal stopping distributions and hazard gradients}
Randomized stopping forms a weighted average of the fixed depth utilities. Its optimum therefore depends on which depths maximize utility, as the following result makes precise.

\begin{theorem}[Existence and support of an optimal stopping distribution]
\label{thm:halt-stationary}
For fixed $W_d(s)$ and $c$, maximize
\begin{equation}
J_s(h)=\sum_{d=1}^{\dmax}q_h(d\mid s)U_d(s;c)
\label{eq:halt-objective}
\end{equation}
over unrestricted hazards $h\in[0,1]^{\dmax-1}$. The maximum is $M=\max_dU_d$. A deterministic distribution at any maximizing depth attains it and a stopping distribution is optimal if and only if it is supported on maximizing depths. Under \Cref{prop:halt-optimal}'s assumption, its first-crossing depth is one such choice.
\end{theorem}
\begin{proof}
\emph{Step 1: bound every randomized choice by the best depth.}
The stopping-law lemma gives $q_h(d\mid s)\ge0$ and $\sum_dq_h(d\mid s)=1$.
Since $M=\max_dU_d$, each difference $M-U_d$ is nonnegative and
\[
M-J_s(h)
=M\sum_dq_h(d\mid s)-\sum_dq_h(d\mid s)U_d
=\sum_dq_h(d\mid s)(M-U_d)\ge0.
\]
Equality holds exactly when every depth given positive probability has $U_d=M$.
This proves the support condition: randomizing over suboptimal depths cannot attain the optimum.

\emph{Step 2: show that the bound is attainable with hazards.}
Choose a maximizing depth $d_0$.
Set all earlier hazards to zero, so the policy certainly reaches it.
If $d_0<\dmax$, set $h_{d_0}=1$; if $d_0=\dmax$, the cap supplies the forced stop.
This realizes $q_h(d_0\mid s)=1$ and attains $M$.
Any hazards after a certain stop have no effect.

\emph{Step 3: verify that mixtures over tied optima are feasible too.}
More generally, let $\nu$ be any distribution over depths and define its remaining mass
$R_d=\sum_{k=d}^{\dmax}\nu(k)$.
For $d<\dmax$ with $R_d>0$, choose $h_d=\nu(d)/R_d$.
Start with $S_1=R_1=1$.
If $S_d=R_d>0$, then
\[
q_h(d)=S_dh_d=\nu(d),\qquad
S_{d+1}=S_d(1-h_d)=R_d-\nu(d)=R_{d+1}.
\]
If $R_d=0$, all later masses are zero and their hazards can be arbitrary.
Induction gives the desired probabilities below the cap and $q_h(\dmax)=R_{\dmax}=\nu(\dmax)$ at the cap.
Every distribution supported on maximizing depths is therefore realizable and optimal.
\end{proof}
A unique maximizing depth gives a unique optimal depth distribution, while hazards after a certain stop remain arbitrary.
Sigmoid heads with finite logits lie in $(0,1)$, so deterministic boundary distributions may only be approached.
Shared network parameters also couple decisions across states and independently optimal depths need not be jointly realizable.
The optimization is over free hazards for a fixed value profile; the jointly parameterized PPO training problem is analyzed separately through its surrogate.

For a non-optimal stopping distribution, the local hazard gradient compares stopping now with the continuation strategy currently in use.

\begin{proposition}[The local hazard compares stopping with continuation]
\label{prop:hazard-gradient}
For $d<\dmax$, let the downstream stopping law after forcing continuation past $d$ be
\[
\widetilde q_{d,k}=h_k\prod_{j=d+1}^{k-1}(1-h_j)\ (d<k<\dmax),
\qquad
\widetilde q_{d,\dmax}=\prod_{j=d+1}^{\dmax-1}(1-h_j),
\]
and define $\overline U_{d+}=\sum_{k=d+1}^{\dmax}\widetilde q_{d,k}U_k$.
Holding the value profile and all other hazards fixed,
\begin{equation}
\frac{\partial J_s}{\partial h_d}=S_d(U_d-\overline U_{d+}).
\label{eq:halt-gradient}
\end{equation}
\end{proposition}
\begin{proof}
\emph{Step 1: isolate decisions affected by $h_d$.}
The total utility from stops before $d$ is
$A_{<d}=\sum_{k<d}q_h(k\mid s)U_k$ and does not depend on $h_d$.
Reaching $d$ has probability $S_d$, which also depends only on earlier hazards.
Once $d$ is reached, the policy stops with probability $h_d$ or continues with probability $1-h_d$.

\emph{Step 2: write the value of continuing.}
After forcing continuation, the downstream hazards define $\widetilde q_{d,k}$ on depths $k>d$.
The stopping-law telescoping calculation shows that these probabilities sum to one.
Their weighted utility is $\overline U_{d+}$.
It uses only later hazards and remains well defined even if $S_d=0$.
Consequently,
\[
J_s(h)=A_{<d}+S_d\big[h_dU_d+(1-h_d)\overline U_{d+}\big].
\]

\emph{Step 3: differentiate this expression.}
Holding the other hazards and all $U_k$ fixed makes $A_{<d}$, $S_d$ and $\overline U_{d+}$ constant with respect to $h_d$.
Thus
\[
\frac{\partial J_s}{\partial h_d}
=S_d\frac{\partial}{\partial h_d}
   [h_dU_d+(1-h_d)\overline U_{d+}]
=S_d(U_d-\overline U_{d+}).
\]
This proves the formula, including boundary hazards as a polynomial derivative.
\end{proof}
Thus the gradient compares current utility with the current downstream strategy, which may differ from both the next depth and the optimal continuation.
If $S_d=0$, changing the hazard cannot affect the objective because that depth is never reached.
For $h_d=\sigma(\ell_d)$, the logit derivative also contains $h_d(1-h_d)$ and can be small near saturation.

\subsection{Compute prices and value-estimation error}
We first vary the cost while holding the value profile fixed, then bound the effect of uniformly inaccurate values.

\begin{proposition}[Optimal depth is non-increasing in its price]
\label{prop:cost-monotonicity}
Fix the value profile. If $0\le c_1<c_2$ and $d_i\in\arg\max_d[W_d-c_id]$, then $d_2\le d_1$, without a diminishing-returns assumption.
\end{proposition}
\begin{proof}
Compare the two chosen depths under each price.
Because $d_1$ is optimal at $c_1$ and $d_2$ is optimal at $c_2$,
\begin{align*}
W_{d_1}-c_1d_1&\ge W_{d_2}-c_1d_2,\\
W_{d_2}-c_2d_2&\ge W_{d_1}-c_2d_1.
\end{align*}
Adding cancels both value terms and leaves
\[
(c_2-c_1)(d_1-d_2)\ge0.
\]
The price difference is strictly positive, so division gives $d_1-d_2\ge0$.
The argument permits ties in either maximization and does not require monotone value increments.
\end{proof}
This comparative statement holds with the action/value profile frozen; retraining at a different $\lambda_c$ can change that profile and need not preserve the relation.

\begin{proposition}[A uniform-error guarantee for value-based selection]
\label{prop:value-error}
Suppose a value estimate $\widetilde W_d$ satisfies $\max_d|\widetilde W_d-W_d|\le\xi$. Let $d^*\in\arg\max_d(W_d-cd)$ and $\widetilde d\in\arg\max_d(\widetilde W_d-cd)$. Then
\begin{equation}
0\le U_{d^*}-U_{\widetilde d}\le2\xi.
\label{eq:value-regret}
\end{equation}
If the true maximizer is unique with gap greater than $2\xi$ to every other depth, the estimated maximizer equals it.
\end{proposition}
\begin{proof}
\emph{Step 1: transfer the value error to utility.}
Both utilities subtract the same exact cost $cd$.
Therefore, for $\widetilde U_d=\widetilde W_d-cd$,
$|\widetilde U_d-U_d|=|\widetilde W_d-W_d|\le\xi$.

\emph{Step 2: insert the estimated utilities into the regret.}
Add and subtract the estimates at the two selected depths:
\begin{align*}
U_{d^*}-U_{\widetilde d}
&=(U_{d^*}-\widetilde U_{d^*})
 +(\widetilde U_{d^*}-\widetilde U_{\widetilde d})
 +(\widetilde U_{\widetilde d}-U_{\widetilde d})\\
&\le\xi+0+\xi=2\xi.
\end{align*}
The middle term is nonpositive because $\widetilde d$ maximizes the estimated utility.
The outer terms are each at most $\xi$ by Step 1.
True optimality of $d^*$ gives the lower bound zero.

\emph{Step 3: use a strict gap to obtain exact selection.}
If every other depth loses more than $2\xi$ in true utility, selecting such a depth would violate Step 2.
Thus a unique optimum with that gap must also be the estimated optimum.
\end{proof}
The corresponding bound for an adjacent increment follows from the triangle inequality:
\[
\big|(\widetilde W_{d+1}-\widetilde W_d)-(W_{d+1}-W_d)\big|
\le|\widetilde W_{d+1}-W_{d+1}|+|\widetilde W_d-W_d|
\le2\xi.
\]
If the true first-crossing rule is valid and every examined increment has margin $|g_d-c|>2\xi$, its threshold decisions are preserved.
This conclusion requires accuracy relative to the counterfactual $W_d$.

\subsection{Discounting and the values required by the analysis}
With discounted returns, $Q^{\bar\pi}$ already includes the environment discount factor. Internal refinement is instantaneous in this model, so no additional $\gamma^d$ factor belongs in \Cref{eq:halt-utility}. Nevertheless, changing $\gamma$ can change the ordering and increments of $W_d$; it does not generally cancel from their differences. If computation consumes environment time or causes delayed actions, that delay requires a different value/cost model.

The first-crossing assumption concerns increments of the counterfactual action values $W_d$.
The shared-target predictions are characterized separately in \Cref{app:shared-target-proof}.

\section{Tasks}
\label{app:tasks-app}
The benchmark combines tasks with different observations, action spaces and opportunities for useful refinement. \Cref{tab:tasks-app} summarizes the twelve-task set; the language task is specified below.

\begin{table}[h]
\centering
\caption{\textbf{Tasks.} Every benchmark setting, including DistShift, uses five seeds; \Cref{app:inventory} summarizes the protocols of the additional studies.}
\label{tab:tasks-app}
\scriptsize
\setlength{\tabcolsep}{4pt}
\begin{tabular}{@{}l l l c@{}}
\toprule
Task & Domain & Source & Role \\
\midrule
\TaskControl  & Continuous control & Gym MuJoCo & anchor \\
\TaskManip    & Robotic manipulation & ManiSkill3 & anchor \\
\TaskStack    & Sequential manipulation & ManiSkill3 & anchor \\
\TaskNav      & Maze navigation & D4RL/Gymnasium-Robotics & anchor \\
\TaskNavNeg   & Maze nav.\ (low-heterogeneity) & D4RL variant & low-het. \\
\TaskLLM      & Entailment classification  & GLUE-RTE on GPT-2 & anchor \\
\TaskMGDist   & Gridworld dist.\ shift & MiniGrid & sat.\ ctrl. \\
\TaskMGFour   & Gridworld FourRooms & MiniGrid & anchor \\
\TaskOGCube   & Goal-cond.\ multi-cube & OGBench & goal-cond. \\
\TaskOGScene  & Goal-cond.\ scene & OGBench & goal-cond. \\
\TaskOGPuzzle & Goal-cond.\ puzzle & OGBench & goal-cond. \\
\TaskRare     & Rare-event manipulation & ManiSkill3 (slip/misalign) & rare \\
\bottomrule
\end{tabular}
\end{table}

\subsection{Language Task: GPT-2 on GLUE-RTE}
\label{app:tasks-llm}

\TaskLLM{} tests adaptive refinement outside sequential control.
The observation is an encoded text pair and the action is a binary label, allowing us to study allocation with a frozen pretrained encoder and a small action space.

\paragraph{Task.}
GLUE-RTE~\citep{wang2018glue} is binary recognizing-textual-entailment: given a (premise, hypothesis) token sequence, predict \texttt{entailment} or \texttt{not\_entailment}.
We use the standard $2{,}490$-example train split and the $277$-example dev split as evaluation; the test split has hidden labels and is not used.
Each example is a single-step episode: the agent observes the tokenized pair, picks a label, receives reward $+1$ if correct and $0$ otherwise and the episode ends.
This collapses RL to bandit-style classification but preserves AIC's compute-allocation question: how many refinement steps to spend on this example before committing to a label.

\paragraph{Encoder.}
The encoder is GPT-2-small with 124M frozen parameters~\citep{radford2019language}; the language-model head is unused and gradients do not update GPT-2.
The GPT-2 BPE tokenizer represents each input as \texttt{<premise> <sep> <hypothesis>}, padded or truncated to 256 tokens.
We take the 768-dimensional hidden state at the final non-padding token of the last layer and project it through a trained linear layer to the 256-dimensional refinement representation.

\paragraph{Refinement and heads.}
The shared MLP applies $z_d=z_{d-1}+\mathrm{MLP}(z_{d-1})$ to the projected representation.
Each depth has a two-class softmax action head $\pi_d$, a scalar value head $V_d$ and a sigmoid halting head $h_d$.
Every value head receives the same GAE target, which reduces here to the one-step Monte Carlo return.
The selected label and stopping depth enter the joint PPO surrogate.

\paragraph{Evaluation and aggregation.}
Each checkpoint is evaluated ten times on the 277-example dev split, with actions sampled from the selected depth's action distribution and its accuracy is the mean over these passes.
Tables report the mean $\pm$ standard deviation of the per-seed accuracies over five seeds.
One dev example corresponds to $1/277\approx.0036$ accuracy, so the $B=8$ gain of $+.0023$ in \Cref{tab:main} amounts to $0.64$ examples per seed.
The confidence-threshold baseline selects its entropy threshold on a held-out 249-example slice of the training split, which is excluded from policy training.

\paragraph{Allocated computation.}
AIC allocates refinement after text encoding: GPT-2's forward pass is unchanged and the task makes one classification decision rather than generating tokens autoregressively.
The benchmark checkpoints use a mean depth of $3.07$; the timing study in \Cref{app:efficiency} uses a separately trained checkpoint at $\lambda_c=.003$ whose mean dev-set depth is $4.08$.

\paragraph{Critical-state annotation.}
The decision unit is the whole premise--hypothesis pair rather than individual tokens, so critical-state annotation labels whole examples.

\subsection{POGP Comparison Environments}
\label{app:pogp}
\Cref{tab:pogp-aic} uses the Gym MuJoCo locomotion tasks HalfCheetah-v4, Walker2d-v4, Ant-v4 and Hopper-v4 with POGP's training and evaluation protocol and unnormalized returns, so these runs are separate from the benchmark's HalfCheetah task.
POGP's protocol evaluates the first 100 environment steps of each episode, so its returns are about a tenth of the usual 1000-step returns and include the acceleration from rest at the start of every episode.
AIC is attached to POGP's diffusion policy as in \Cref{app:transfer}: its halting head reads the denoiser's hidden features, one iteration is one denoising step in both methods and both methods share the eight-step cap.

\section{Relation to ACT, PonderNet and Metacontrol}
\label{app:precursor}
\citet{ghugare2026computation} study fixed-depth iterative policies and identify adaptive depth within one policy as an open direction.
AIC learns that decision using established stopping and metacontrol ideas.

\paragraph{Established stopping mechanisms.}
ACT makes recurrent computation input-dependent through a task objective and a ponder cost~\citep{graves2016adaptive}.
PonderNet learns a distribution over iterative prediction depths~\citep{banino2021pondernet}, while metacontrol uses RL to choose the identity and number of internal expert queries~\citep{hamrick2017metacontrol}.
Adaptive depth, probabilistic stopping and reward-sensitive computation therefore precede AIC.

\paragraph{Policy formulation and optimization.}
ACT combines intermediate outputs using accumulated halting mass and a ponder penalty.
PonderNet uses conditional stopping probabilities, an expected prediction loss and a KL penalty toward a geometric stopping prior.
AIC uses the same general conditional-stopping construction to select one depth-specific action policy.
Its joint action/stopping probability enters the PPO ratio, with an expected-depth penalty and auxiliary value supervision (\Cref{sec:method-voc}).
Earlier metacontrol also learns computation decisions through RL; AIC applies stopping directly inside the action policy's refinement process.
A constant cost permits state-dependent stopping in both settings (\Cref{app:voc-vs-ponder}).

\paragraph{Allocation analysis and empirical attribution.}
\Cref{thm:main} compares an ideal state-dependent allocator with state-independent stopping at equal expected depth, under a fixed state distribution and explicit value-profile assumptions.
The experiments separate the effects of stopping, supervision, data collection and refinement from the narrower effect of reassigning equal computation across states and \Cref{app:supervision-vs-allocation} analyzes the task-specific factorial effects.

\section{Factorial Details}
\label{app:supervision-vs-allocation}
The four configurations in \Cref{tab:alloc-vs-sup} vary adaptive stopping and auxiliary per-depth value supervision.
The auxiliary per-depth loss is distinct from the ordinary value baseline used to estimate PPO advantages and every configuration is an independently trained configuration.

\paragraph{Conditional effects and interaction.}
For $R_{as}$ as defined in \Cref{sec:factorial}, the interaction is
\begin{equation}
I=(R_{11}-R_{01})-(R_{10}-R_{00})
  =(R_{11}-R_{10})-(R_{01}-R_{00}).
\end{equation}
From the configuration means, $I=.003$ for PickCube, $.006$ for AntMaze, $.004$ for Memory, $.002$ for DoorKey, $-.005$ for StackCube and zero for RareEvent.
These interactions are small relative to the main effects but not uniformly zero, so we give both conditional effects rather than assume additive effects.

\paragraph{From configuration changes to allocation.}
An adaptive setting can change mean depth, the distribution of depths, the action heads learned during training and which states receive computation.
The stopping effect therefore combines these changes.
In our factorial the non-adaptive configurations execute one refinement per decision and the adaptive configurations execute 1.8--3.9 on average (\Cref{tab:alloc-vs-sup}), so each stopping effect includes additional computation as well as state-dependent assignment.
An allocation-specific comparison would replace the non-adaptive schedule with a separately tuned state-independent schedule at the adaptive configuration's mean computation, keeping the encoder, refinement cost and training update budget fixed.
Changing the supervision factor preserves the ordinary advantage baseline and the parameter count, since the value heads remain in the network and only their loss weight is set to zero.
Treating configurations as independent, the PickCube stopping effects have standard errors of $.0032$ and $.0029$ and its supervision effects $.0035$ and $.0025$; the RareEvent stopping effects of $.002$ lie within one standard error, whereas its supervision effects of $.031$ exceed six.

\section{Halting Baselines}
\label{app:adaptive-baselines-extra}
\label{app:baseline-defs}
\paragraph{Untrained stopping rules.}
The confidence rule stops when policy entropy falls below a validation-selected threshold.
Action agreement compares adjacent action distributions, while the value-gain rule compares adjacent value predictions; both need a previous depth, so their earliest stop is depth two.
A mean depth of exactly $2.00$ in \Cref{tab:halt-step-app} therefore means that the rule stopped at depth two on every decision.
This holds for both rules on DistShift and RareEvent, where the two rules coincide; their RareEvent reward of $.4198$ measures the depth-two action head on every state.
The random schedule stops at each depth $d<\dmax$ with a state-independent probability $\beta$, so $q(d)=\beta(1-\beta)^{d-1}$ for $d<\dmax$, the cap receives $(1-\beta)^{\dmax-1}$ and $\E[\tau]=(1-(1-\beta)^{\dmax})/\beta$.
For each task, $\beta$ maximizes validation return under a golden-section search over $\log\beta\in[\log10^{-3},0]$; \Cref{tab:halt-step-app} lists the selected values.
The schedule is therefore a tuned state-independent stopping law, the empirical counterpart of the baseline class in \Cref{thm:main}, without matching AIC's mean depth.
With $\beta=1$ it executes one refinement on every decision, as on DistShift, low-heterogeneity AntMaze and RareEvent.
All untrained rules are applied to the trained AIC network at $B=8$ and replace only its halting head.

\paragraph{Jointly trained baselines.}
\Cref{tab:halt-spec} specifies the two jointly trained baselines alongside AIC.
ACT-ponder follows \citet{graves2016adaptive}: its halting units accumulate halting mass until it exceeds $1-\varepsilon_{\mathrm{ACT}}$ with $\varepsilon_{\mathrm{ACT}}=.01$, the action distribution is the halting-weighted mixture of the depth-specific heads and the ponder cost $N+R$ enters the PPO loss with weight $.01$.
The learned non-VoC baseline keeps AIC's architecture and evaluation rule and changes only the halting objective: its stopping probabilities are excluded from the PPO ratio and a two-layer halting MLP reads a stop-gradient copy of $z_d$ and is trained by binary cross-entropy toward $\mathbf 1\{\mathrm{KL}(\pi_d\,\|\,\pi_{d+1})<.01\}$ on the same rollout states.
The name records that this stopping signal is learned without value-of-computation credit: no return or advantage reaches its halting decision.
All three methods can vary depth across states, including with a constant per-step cost and we evaluate their recorded reward and mean depth together.

\begin{table}[h]
\centering
\caption{\textbf{Jointly trained halting methods.} All three share the encoder, refinement block, depth-specific action heads and value heads and use five seeds at $B=8$.}
\label{tab:halt-spec}
\scriptsize
\setlength{\tabcolsep}{3pt}
\begin{tabular}{@{}P{.19\linewidth}P{.25\linewidth}P{.25\linewidth}P{.25\linewidth}@{}}
\toprule
& AIC & ACT-ponder & Learned non-VoC\\
\midrule
Halting input & $z_d$ & $z_d$ & stop-gradient copy of $z_d$\\
Halting head & linear, one per depth $d<\dmax$ & linear, shared across depths & two-layer MLP ($256{\to}160{\to}1$), shared\\
Halting objective & joint PPO ratio over $(\tau,a)$ & task loss through the halting-weighted action mixture & cross-entropy toward $\mathbf 1\{\mathrm{KL}(\pi_d\|\pi_{d+1})<.01\}$\\
Stopping in PPO ratio & yes & through the mixture likelihood & no\\
Computation penalty & $\lambda_c\E[\tau]/\dmax$, $\lambda_c=.01$ & ponder cost $N+R$, weight $.01$ & none\\
Per-depth value supervision & yes, $\lambda_V=.5$ & yes, $\lambda_V=.5$ & yes, $\lambda_V=.5$\\
Evaluation rule & first $h_d>.5$, else cap & halt once accumulated mass exceeds $1-\varepsilon_{\mathrm{ACT}}$ & first $h_d>.5$, else cap\\
Parameters & 2.071M & 2.069M & 2.110M\\
\bottomrule
\end{tabular}
\end{table}

\paragraph{Task-dependent tradeoffs.}
RareEvent has small truncation effects and larger supervision effects.
On FourRooms, ACT has both greater mean depth and higher reward than AIC, placing it at a different return--compute operating point.

\section{Refinement Depth and Computation Accounting}
\label{app:halt-step-full}
\label{app:depth-dist}

Mean depth measures how many refinement steps a policy executes per decision.
\Cref{tab:halt-step-app} compares that quantity across the twelve benchmark tasks.
End-to-end timing additionally includes the encoder and control-flow overhead (\Cref{app:efficiency}).

\begin{table}[h]
\centering
\caption{\textbf{Mean refinement depth at $B{=}8$ on all twelve tasks.}
Mean steps per state for the seven stopping methods; trained-method entries carry the seed standard deviation as a subscript and $\beta$ is the random schedule's per-task stopping probability (\Cref{app:baseline-defs}). Fixed-$K$ controls execute exactly $K$ refinements. ACT and non-VoC halt deeper than AIC on eleven tasks; HalfCheetah is the exception. The final row is the unweighted task mean.}
\label{tab:halt-step-app}
\scriptsize
\setlength{\tabcolsep}{3.5pt}
\begin{tabular}{@{}l ccc cccc c@{}}
\toprule
& \multicolumn{3}{c}{Trained} & \multicolumn{4}{c}{Untrained / heuristic} & \\
\cmidrule(lr){2-4}\cmidrule(lr){5-8}
Task & AIC & ACT & non-VoC & Conf. & Agr. & Val.-g. & Rand. & $\beta$ \\
\midrule
HalfCheetah-v4         & $3.04_{\pm.07}$ & $2.94_{\pm.08}$ & $2.93_{\pm.08}$ & $2.90$ & $2.85$ & $3.24$ & $2.00$ & $.498$ \\
ManiSkill PickCube-v1  & $3.47_{\pm.09}$ & $3.80_{\pm.14}$ & $3.85_{\pm.12}$ & $3.51$ & $3.79$ & $3.38$ & $3.11$ & $.304$ \\
ManiSkill StackCube-v1 & $3.77_{\pm.11}$ & $4.80_{\pm.17}$ & $4.75_{\pm.15}$ & $3.86$ & $3.42$ & $3.86$ & $3.55$ & $.255$ \\
AntMaze (main)         & $3.15_{\pm.08}$ & $3.54_{\pm.12}$ & $3.65_{\pm.13}$ & $3.07$ & $2.85$ & $3.14$ & $2.12$ & $.469$ \\
AntMaze (low-het.)     & $2.24_{\pm.06}$ & $2.64_{\pm.09}$ & $2.56_{\pm.08}$ & $2.16$ & $2.22$ & $2.22$ & $1.00$ & $1$ \\
GPT-2 on GLUE-RTE      & $3.07_{\pm.06}$ & $3.42_{\pm.09}$ & $3.33_{\pm.08}$ & $3.23$ & $3.04$ & $2.93$ & $1.30$ & $.769$ \\
MiniGrid-DistShift2    & $1.00_{\pm.00}$ & $1.04_{\pm.03}$ & $1.01_{\pm.01}$ & $1.00$ & $2.00$ & $2.00$ & $1.00$ & $1$ \\
MiniGrid-FourRooms     & $2.43_{\pm.12}$ & $6.82_{\pm.21}$ & $6.19_{\pm.19}$ & $2.28$ & $2.46$ & $2.26$ & $1.35$ & $.741$ \\
OG cube-double-play    & $5.19_{\pm.14}$ & $6.68_{\pm.19}$ & $6.67_{\pm.17}$ & $5.27$ & $2.84$ & $4.93$ & $7.59$ & $.0151$ \\
OG scene-play          & $5.05_{\pm.13}$ & $6.23_{\pm.18}$ & $6.22_{\pm.16}$ & $4.98$ & $2.78$ & $5.13$ & $7.19$ & $.0307$ \\
OG puzzle-4x4-play     & $5.60_{\pm.15}$ & $7.20_{\pm.16}$ & $7.26_{\pm.15}$ & $5.64$ & $2.49$ & $5.26$ & $7.92$ & $.0029$ \\
RareEvent-Misalign     & $1.86_{\pm.05}$ & $1.99_{\pm.04}$ & $1.99_{\pm.04}$ & $1.50$ & $2.00$ & $2.00$ & $1.00$ & $1$ \\
\midrule
Mean (12 tasks)        & $3.32$ & $4.26$ & $4.20$ & $3.28$ & $2.73$ & $3.36$ & $3.26$ & --- \\
\bottomrule
\end{tabular}
\end{table}

\Cref{tab:halt-step-app} gives average refinement use; \Cref{app:critstate,app:semantic} examine how depth is distributed across states.

\paragraph{Combining depth with reward.}
\Cref{fig:depth-bars} combines the mean depths with the seven-task rewards in \Cref{tab:baselines-adaptive}.
A fixed baseline's depth is the seed average of its executed depths $\min(K_{\rm train},8)$.
For AntMaze and RareEvent, the fixed rewards pool training depths $2$ and $8$ (one and four seeds) and $1$ and $8$ (two and three seeds), giving mean executed depths of $6.8$ and $5.2$.
On FourRooms, AIC's lower mean depth accompanies lower mean reward than the trained alternatives.

\section{Hyperparameters and Sensitivity on Failure Tasks}
\label{app:hyperparams}
We vary the computation and value-loss weights to test whether the two default-setting failures persist.

\paragraph{Default configuration.}
The benchmark draws on MuJoCo/Gym~\citep{brockman2016openai,todorov2012mujoco}, ManiSkill3~\citep{tao2025maniskill3gpuparallelizedrobotics}, D4RL-style navigation~\citep{fu2020d4rl}, MiniGrid~\citep{chevalier2018minimalistic}, OGBench~\citep{park2024ogbench} and GLUE-RTE~\citep{wang2018glue}.
The twelve-task benchmark uses Adam, learning rate $3\times10^{-4}$, $\gamma=.99$, GAE $\lambda_{\rm GAE}=.95$, PPO clip $\eta=.2$, ten PPO epochs, batch size 64, four parallel environments, hidden dimension 256, $\dmax=8$, $\lambda_c=.01$ and $\lambda_V=.5$.
Anchor tasks use seeds 41--45 and DistShift also uses five seeds.
The two loss weights were selected on PickCube and shared across all twelve benchmark tasks.
$\lambda_c$ is expressed relative to the extrinsic-reward scale; a single numerical value therefore need not represent the same tradeoff across tasks.
Reward normalization and a dual variable for an average-compute constraint are possible extensions.

\paragraph{Loss-weight selection on PickCube.}
Sweeping the compute weight over $\{0,.001,.005,.01,.05,.1\}$ gives PickCube rewards of $.4897$--$.4904$ and sweeping the value weight over $\{0,.1,.5,1,2\}$ gives $.4898$--$.4925$, so PickCube is insensitive to both weights.
We then test whether these weights transfer to FourRooms and Puzzle.

\paragraph{Failure-task sweeps.}
We sweep $\lambda_c\in\{.001,.003,.01,.03,.1\}$ and $\lambda_V\in\{.25,.5,1,2\}$ on FourRooms and Puzzle, comparing each setting with a fixed-depth baseline matched to AIC's mean inference depth.
\begin{table}[h]
\centering\small
\caption{FourRooms: difference from the mean-depth-matched fixed baseline in percentage points of normalized reward (one point is $.01$ in the units of \Cref{tab:main}), mean $\pm$ standard error over five seeds.}
\label{tab:fourrooms-sweep}
\begin{tabular}{lrrrr}
\toprule
$\lambda_c$ & $\lambda_V=.25$ & $\lambda_V=.5$ & $\lambda_V=1$ & $\lambda_V=2$\\
\midrule
.001 & $-.9\pm.3$ & $-.7\pm.3$ & $-.8\pm.4$ & $-1.0\pm.4$\\
.003 & $-1.0\pm.3$ & $-.9\pm.3$ & $-1.0\pm.3$ & $-1.2\pm.4$\\
.010 & $-1.7\pm.4$ & $-1.8\pm.4$ & $-1.6\pm.4$ & $-1.8\pm.5$\\
.030 & $-1.8\pm.4$ & $-1.9\pm.4$ & $-1.8\pm.5$ & $-2.0\pm.5$\\
.100 & $-2.0\pm.5$ & $-2.1\pm.5$ & $-2.0\pm.5$ & $-2.1\pm.5$\\
\bottomrule
\end{tabular}
\end{table}
FourRooms stays negative across the grid; tuning narrows its default deficit from $-1.8\pm.4$ to $-.7\pm.3$.
On Puzzle (\Cref{tab:puzzle-sweep}), the differences are positive, $+.6$ to $+.9$ percentage points, for every $\lambda_c$ when $\lambda_V\in\{1,2\}$ and negative for $\lambda_V\le.5$, including $-.5\pm.3$ at the default setting.

\begin{table}[h]
\centering\small
\caption{Puzzle: difference from the mean-depth-matched fixed baseline in percentage points of normalized reward, mean $\pm$ standard error over five seeds.}
\label{tab:puzzle-sweep}
\begin{tabular}{lrrrr}
\toprule
$\lambda_c$ & $\lambda_V=.25$ & $\lambda_V=.5$ & $\lambda_V=1$ & $\lambda_V=2$\\
\midrule
.001 & $-.6\pm.3$ & $-.3\pm.3$ & $+.8\pm.3$ & $+.9\pm.3$\\
.003 & $-.5\pm.3$ & $-.3\pm.3$ & $+.9\pm.3$ & $+.8\pm.3$\\
.010 & $-.7\pm.3$ & $-.5\pm.3$ & $+.7\pm.3$ & $+.8\pm.3$\\
.030 & $-.8\pm.4$ & $-.6\pm.3$ & $+.7\pm.3$ & $+.7\pm.4$\\
.100 & $-1.0\pm.4$ & $-.8\pm.4$ & $+.6\pm.4$ & $+.6\pm.4$\\
\bottomrule
\end{tabular}
\end{table}
Puzzle is thus sensitive to the value weight, while FourRooms is a persistent limitation within the tested range.

\section{Privileged Reference and Remaining Headroom}
\label{app:oracle}
The privileged reference estimates how much performance remains available when depth selection can use task-specific information.
It selects $\min(\hat\kappa(s),B)$ using torque percentiles for HalfCheetah; contact and phase indicators for manipulation; intersection proximity for AntMaze; a label-margin probe for RTE; a slip indicator for RareEvent; and goal-distance bins for OGBench.
This is a heuristic reference under a common depth cap.
Its gap recovery measures progress toward that reference, rather than an optimal-policy bound or a matched-work comparison.
The reference runs AIC's trained network at each budget and changes only the stopping rule, so at $B=1$ it coincides with AIC.
Gap recovery is measured relative to this reference and is undefined when its reward does not exceed the fixed baseline's reward; \Cref{tab:full-budgets} lists the budget-specific reference rewards.

\begin{table}[h]
\centering\small
\caption{Headroom diagnostic on PickCube, AntMaze, Memory and DoorKey: fraction of the privileged-reference gap recovered.}
\label{tab:headroom-probe}
\begin{tabular}{lr}
\toprule
Setting & Gap recovered\\
\midrule
Default halting entropy & 6.1\%\\
$2\times$ halting entropy & 6.8\%\\
$4\times$ halting entropy & 7.4\%\\
Privileged future-benefit probe & 29.8\%\\
\bottomrule
\end{tabular}
\end{table}
The future-benefit probe uses realized future benefit, which is unavailable at deployment; its $29.8\%$, against at most $7.4\%$ from higher halting entropy, points to predicting useful refinement from deployment-time information as the main opportunity.
On an eight-task subset evaluated separately, AIC recovers $5.4\%$ of the gap on average.

\paragraph{Localized refinement on AntMaze.}
With five fresh seeds, AIC recovers $6.3\%$ of the AntMaze reference gap, while the best local-mask variant, which restricts refinement spatially, recovers $14.2\%$ at the same mean depth and improves normalized reward by $2.6$ percentage points; these runs are independent of the $8.1\%$ recovery in \Cref{tab:main}.
Spatially restricted refinement therefore appears to reduce interference from irrelevant features.

\section{Full Nominal-Budget Comparison (12 tasks)}
\label{app:matched-full}
\Cref{tab:full-budgets} gives every nominal-budget comparison in the twelve-task benchmark. The cap bounds the available depth, so mean computation can still differ between methods; \Cref{tab:aic-depth} gives AIC's mean depth at each budget.
For a fixed checkpoint trained at depth $K_{\rm train}$, evaluation executes $K_{\rm exec}=\min(K_{\rm train},B)$ refinements and uses the action head at that executed depth.
The parenthesized depths identify checkpoint training depths selected across seeds, with superscripts giving the number of seeds that selected each depth when more than one was selected.
Thus $(1^{(4)},4^{(1)})$ at $B=1$ means that four seeds selected a depth-one checkpoint and one seed a depth-four checkpoint, both executing one refinement; $(4^{(3)},8^{(2)})$ at $B=4$ similarly executes four.
Each budget's baselines come from that budget's training sweep (\Cref{sec:matched}), so a depth-one checkpoint selected at $B=8$ is a different training run from the depth-one checkpoints at $B=1$.
Derived differences, relative gains and reference-gap recoveries in \Cref{tab:main,tab:full-budgets} are computed from the displayed means, before rounding the derived quantity; each budget uses its own reference reward, listed in the Ref.\ column.
Against the strongest fixed baseline at any budget up to $B$, AIC has 6, 8 and 7 positive differences at $B=2$, 4 and 8, compared with 8, 10 and 9 against the per-budget baselines; low-heterogeneity AntMaze and RareEvent account for the difference at every budget because their depth-one baselines lead throughout.

\begin{table}[h]
\centering
\scriptsize
\setlength{\tabcolsep}{3pt}
\caption{Full comparison at nominal $B \in \{1,2,4,8\}$.
Means $\pm$ sample standard deviation over 5 seeds.
Parentheses list training depths selected across seeds, with seed counts as superscripts when more than one depth was selected; executed depth is $\min(K_{\rm train},B)$.
Ref.: reward of the budget-specific privileged reference. Gap recovery is undefined when the reference does not exceed the fixed baseline and is omitted at $B=1$, where every method executes one refinement.
\TaskMGDist{} is the saturated control.
AIC trails on \TaskMGFour{} and \TaskOGPuzzle{} at $B{=}8$.}
\label{tab:full-budgets}
\begin{tabular}{@{}l c cc c c c c@{}}
\toprule
Task & $B$ & AIC & Fixed ($K_{\rm train}$) & $\Delta$ & Ref. & Ref. gap & W/L/T \\
\midrule
\multirow{4}{*}{\TaskMGDist}
& 1 & $1.0000{\pm}.0000$ & $1.0000{\pm}.0000$ ($1$) & $.0000$ & --- & --- & 0/0/5 \\
& 2 & $1.0000{\pm}.0000$ & $1.0000{\pm}.0000$ ($1$) & $.0000$ & $1.0000$ & --- & 0/0/5 \\
& 4 & $1.0000{\pm}.0000$ & $1.0000{\pm}.0000$ ($1$) & $.0000$ & $1.0000$ & --- & 0/0/5 \\
& 8 & $1.0000{\pm}.0000$ & $1.0000{\pm}.0000$ ($1$) & $.0000$ & $1.0000$ & --- & 0/0/5 \\
\midrule
\multirow{4}{*}{\TaskControl}
& 1 & $.9073{\pm}.0008$ & $.9079{\pm}.0008$ ($1$) & $-.0006$ & --- & --- & 2/3/0 \\
& 2 & $.9269{\pm}.0006$ & $.9270{\pm}.0003$ ($2$) & $-.0001$ & $.9437$ & $-0.6\%$ & 3/2/0 \\
& 4 & $.9331{\pm}.0004$ & $.9314{\pm}.0002$ ($4$) & $+.0017$ & $.9612$ & $5.7\%$ & 5/0/0 \\
& 8 & $.9348{\pm}.0001$ & $.9329{\pm}.0002$ ($8$) & $+.0019$ & $.9744$ & $4.6\%$ & 5/0/0 \\
\midrule
\multirow{4}{*}{\TaskLLM}
& 1 & $.7912{\pm}.0027$ & $.7932{\pm}.0022$ ($1^{(4)},4^{(1)}$) & $-.0020$ & --- & --- & 2/2/1 \\
& 2 & $.8230{\pm}.0042$ & $.8216{\pm}.0020$ ($2^{(3)},4^{(2)}$) & $+.0014$ & $.8852$ & $2.2\%$ & 3/2/0 \\
& 4 & $.8298{\pm}.0011$ & $.8276{\pm}.0042$ ($4$) & $+.0022$ & $.9193$ & $2.4\%$ & 3/1/1 \\
& 8 & $.8334{\pm}.0009$ & $.8311{\pm}.0024$ ($8$) & $+.0023$ & $.9323$ & $2.3\%$ & 4/1/0 \\
\midrule
\multirow{4}{*}{\TaskOGCube}
& 1 & $.2453{\pm}.0019$ & $.2446{\pm}.0031$ ($1$) & $+.0007$ & --- & --- & 3/2/0 \\
& 2 & $.3311{\pm}.0036$ & $.3305{\pm}.0056$ ($2$) & $+.0006$ & $.2555$ & --- & 3/2/0 \\
& 4 & $.4047{\pm}.0060$ & $.4033{\pm}.0047$ ($4$) & $+.0014$ & $.3871$ & --- & 3/2/0 \\
& 8 & $.4299{\pm}.0028$ & $.4281{\pm}.0060$ ($8$) & $+.0018$ & $.5646$ & $1.3\%$ & 4/1/0 \\
\midrule
\multirow{4}{*}{\TaskOGScene}
& 1 & $.2535{\pm}.0042$ & $.2538{\pm}.0027$ ($1$) & $-.0003$ & --- & --- & 3/2/0 \\
& 2 & $.3216{\pm}.0034$ & $.3152{\pm}.0036$ ($2$) & $+.0064$ & $.3587$ & $14.7\%$ & 5/0/0 \\
& 4 & $.3965{\pm}.0045$ & $.3957{\pm}.0069$ ($4$) & $+.0008$ & $.5957$ & $0.4\%$ & 2/3/0 \\
& 8 & $.4167{\pm}.0048$ & $.4075{\pm}.0044$ ($8$) & $+.0092$ & $.5565$ & $6.2\%$ & 5/0/0 \\
\midrule
\multirow{4}{*}{\TaskManip}
& 1 & $.4062{\pm}.0058$ & $.4132{\pm}.0027$ ($1$) & $-.0070$ & --- & --- & 1/4/0 \\
& 2 & $.4756{\pm}.0018$ & $.4732{\pm}.0051$ ($2$) & $+.0024$ & $.5589$ & $2.8\%$ & 4/1/0 \\
& 4 & $.4954{\pm}.0036$ & $.4911{\pm}.0027$ ($4$) & $+.0043$ & $.6344$ & $3.0\%$ & 5/0/0 \\
& 8 & $.4926{\pm}.0045$ & $.4741{\pm}.0034$ ($8$) & $+.0185$ & $.6756$ & $9.2\%$ & 5/0/0 \\
\midrule
\multirow{4}{*}{\TaskStack}
& 1 & $.3748{\pm}.0060$ & $.3746{\pm}.0037$ ($1$) & $+.0002$ & --- & --- & 2/3/0 \\
& 2 & $.4251{\pm}.0014$ & $.4217{\pm}.0031$ ($2$) & $+.0034$ & $.4884$ & $5.1\%$ & 4/1/0 \\
& 4 & $.4554{\pm}.0024$ & $.4508{\pm}.0071$ ($4$) & $+.0046$ & $.5946$ & $3.2\%$ & 3/2/0 \\
& 8 & $.4579{\pm}.0036$ & $.4458{\pm}.0060$ ($8$) & $+.0121$ & $.6274$ & $6.7\%$ & 4/0/1 \\
\midrule
\multirow{4}{*}{\TaskNav}
& 1 & $.4567{\pm}.0024$ & $.4589{\pm}.0019$ ($1$) & $-.0022$ & --- & --- & 1/4/0 \\
& 2 & $.5017{\pm}.0034$ & $.5040{\pm}.0046$ ($2$) & $-.0023$ & $.6135$ & $-2.1\%$ & 2/3/0 \\
& 4 & $.5208{\pm}.0026$ & $.5095{\pm}.0052$ ($4$) & $+.0113$ & $.7043$ & $5.8\%$ & 4/1/0 \\
& 8 & $.5209{\pm}.0034$ & $.5018{\pm}.0022$ ($2^{(1)},8^{(4)}$) & $+.0191$ & $.7367$ & $8.1\%$ & 5/0/0 \\
\midrule
\multirow{4}{*}{\TaskNavNeg}
& 1 & $.6881{\pm}.0040$ & $.6916{\pm}.0021$ ($1$) & $-.0035$ & --- & --- & 1/4/0 \\
& 2 & $.6752{\pm}.0056$ & $.6630{\pm}.0088$ ($1^{(2)},2^{(3)}$) & $+.0122$ & $.8049$ & $8.6\%$ & 5/0/0 \\
& 4 & $.6743{\pm}.0053$ & $.6683{\pm}.0085$ ($4$) & $+.0060$ & $.8618$ & $3.1\%$ & 4/1/0 \\
& 8 & $.6776{\pm}.0025$ & $.6704{\pm}.0047$ ($8$) & $+.0072$ & $.8685$ & $3.6\%$ & 5/0/0 \\
\midrule
\multirow{4}{*}{\TaskRare}
& 1 & $.6130{\pm}.0052$ & $.6338{\pm}.0039$ ($1$) & $-.0208$ & --- & --- & 0/5/0 \\
& 2 & $.5831{\pm}.0090$ & $.5741{\pm}.0063$ ($2$) & $+.0090$ & $.7084$ & $6.7\%$ & 4/1/0 \\
& 4 & $.5851{\pm}.0151$ & $.5632{\pm}.0064$ ($4$) & $+.0219$ & $.7412$ & $12.3\%$ & 5/0/0 \\
& 8 & $.5918{\pm}.0050$ & $.5575{\pm}.0125$ ($1^{(2)},8^{(3)}$) & $+.0343$ & $.7523$ & $17.6\%$ & 5/0/0 \\
\midrule
\multirow{4}{*}{\TaskMGFour}
& 1 & $.8216{\pm}.0148$ & $.7928{\pm}.0604$ ($1^{(3)},4^{(2)}$) & $+.0288$ & --- & --- & 3/2/0 \\
& 2 & $.8162{\pm}.0103$ & $.8216{\pm}.0040$ ($1^{(2)},2^{(3)}$) & $-.0054$ & $.8154$ & --- & 1/2/2 \\
& 4 & $.9171{\pm}.0040$ & $.9189{\pm}.0000$ ($4^{(3)},8^{(2)}$) & $-.0018$ & $.9153$ & --- & 0/1/4 \\
& 8 & $.9514{\pm}.0103$ & $.9730{\pm}.0064$ ($8$) & $-.0216$ & $.9730$ & --- & 0/5/0 \\
\midrule
\multirow{4}{*}{\TaskOGPuzzle}
& 1 & $.2067{\pm}.0021$ & $.2044{\pm}.0035$ ($1$) & $+.0023$ & --- & --- & 4/1/0 \\
& 2 & $.2997{\pm}.0033$ & $.2909{\pm}.0021$ ($2$) & $+.0088$ & $.3163$ & $34.6\%$ & 5/0/0 \\
& 4 & $.3670{\pm}.0040$ & $.3578{\pm}.0061$ ($4$) & $+.0092$ & $.4235$ & $14.0\%$ & 5/0/0 \\
& 8 & $.4022{\pm}.0048$ & $.4069{\pm}.0025$ ($8$) & $-.0047$ & $.5185$ & $-4.2\%$ & 0/5/0 \\
\bottomrule
\end{tabular}
\end{table}

\begin{table}[h]
\centering
\caption{\textbf{AIC's mean refinement depth at each nominal budget.} $B=1$ forces one refinement; the $B=8$ column repeats \Cref{tab:halt-step-app}.}
\label{tab:aic-depth}
\scriptsize
\setlength{\tabcolsep}{6pt}
\begin{tabular}{@{}l cccc@{}}
\toprule
Task & $B=1$ & $B=2$ & $B=4$ & $B=8$ \\
\midrule
\TaskMGDist & $1.00$ & $1.00$ & $1.00$ & $1.00$ \\
\TaskControl & $1.00$ & $1.83$ & $2.71$ & $3.04$ \\
\TaskLLM & $1.00$ & $1.74$ & $2.62$ & $3.07$ \\
\TaskOGCube & $1.00$ & $1.92$ & $3.58$ & $5.19$ \\
\TaskNavNeg & $1.00$ & $1.52$ & $2.03$ & $2.24$ \\
\TaskOGScene & $1.00$ & $1.90$ & $3.49$ & $5.05$ \\
\TaskStack & $1.00$ & $1.88$ & $3.12$ & $3.77$ \\
\TaskManip & $1.00$ & $1.81$ & $2.93$ & $3.47$ \\
\TaskNav & $1.00$ & $1.79$ & $2.80$ & $3.15$ \\
\TaskRare & $1.00$ & $1.48$ & $1.71$ & $1.86$ \\
\TaskMGFour & $1.00$ & $1.63$ & $2.21$ & $2.43$ \\
\TaskOGPuzzle & $1.00$ & $1.94$ & $3.66$ & $5.60$ \\
\midrule
Mean (12 tasks) & $1.00$ & $1.70$ & $2.66$ & $3.32$ \\
\bottomrule
\end{tabular}
\end{table}

\paragraph{Puzzle across budgets.}
At the default loss weights, AIC improves on Puzzle at $B\in\{1,2,4\}$ and trails at $B=8$; a larger value weight reverses the $B=8$ loss (\Cref{app:hyperparams}), so the budget pattern reflects the training configuration as well as adaptive depth.

\begin{figure}[t]
\centering
\includegraphics[width=\linewidth]{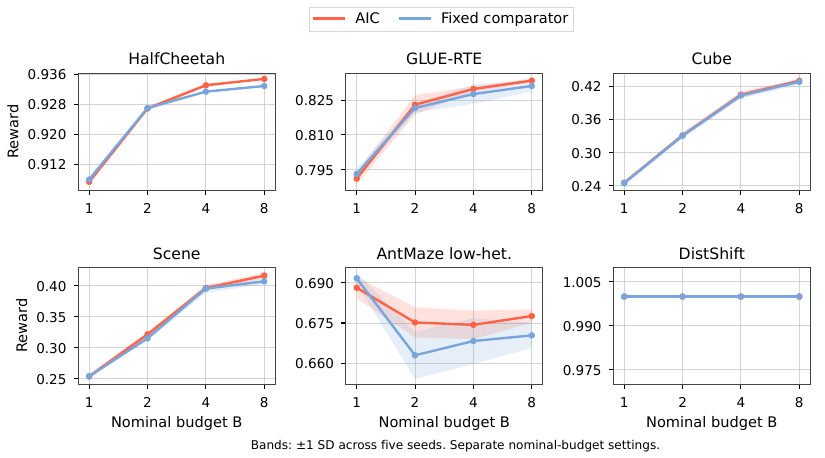}
\caption{\textbf{Performance across nominal budgets on the remaining six benchmark tasks.} Points are five-seed means and shaded bands are $\pm1$ standard deviation from \Cref{tab:full-budgets}. The budget convention and fixed-baseline definitions match \Cref{fig:budget-main}. DistShift is saturated at every budget; the other tasks respond differently to a larger budget.}
\label{fig:pareto}
\end{figure}

\section{Critical-State Methodology}
\label{app:critstate}

\paragraph{Selecting states for early exit.}
For each AIC seed trained at $B=\dmax=8$, we run $N$ evaluation episodes and record the selected depth $\tau_t$ at each visited state.
With $u_{.75}$ the seed's 75th-percentile depth, we re-evaluate the policy with every decision satisfying $\tau_t\ge u_{.75}$ forced to depth one.
Ties at the cutoff can select more than a quarter of decisions, so \Cref{tab:equal-work} lists the realized selected fractions.

\paragraph{Random-state control.}
For each seed, we force a uniformly sampled set of decisions of the same size to depth one and average the drop over ten samples.
This control matches the number of interventions but not the amount of removed computation.

\paragraph{Estimand.}
The drop is $R(\pi)-R(\pi_A)$, where $R(\pi)$ is the return of the unmodified policy and $R(\pi_A)$ the return after applying an early-exit rule $A$.
Because the modified policies can visit different states after their first altered action, targeted-minus-random includes downstream effects on action quality and occupancy.

\paragraph{Equal-work control.}
On each seed's reference set, we record the unmodified depths $\tau_i$ and the removed work $M(A)=\sum_{i\in A}(\tau_i-1)$ of the targeted set.
The equal-work baseline draws decisions uniformly at random, without replacement and independently of task annotations or estimated difficulty, until the cumulative removed work reaches $M(A)$; the last drawn decision is truncated to the intermediate depth that makes the totals equal.
We average its drop over ten draws per seed and on complete rollouts the achieved mean work of the two intervened policies differs by at most 3\%.
This control matches removed computation rather than the number of interventions; \Cref{tab:equal-work} gives the realized selected fractions, the removed refinements per intervened decision and the drops.

\begin{table}[h]
\centering
\caption{\textbf{Removed-work accounting for the truncation study.} Selected fraction: share of decisions at or above the seed's 75th-percentile depth, including ties. Removed refinements are per intervened decision. Equal-work random: random truncation matched to the targeted set's total removed refinements. Targeted drops are those of \Cref{tab:critical-ablation}.}
\label{tab:equal-work}
\scriptsize
\setlength{\tabcolsep}{4pt}
\begin{tabular}{@{}l c cc c c@{}}
\toprule
& Selected & \multicolumn{2}{c}{Removed refinements} & Equal-work & Targeted$-$\\
\cmidrule(lr){3-4}
Task & fraction & Targeted & Random & random drop & equal-work random\\
\midrule
\TaskMGDist\,\scriptsize(sat.) & $1.00$ & $0.00$ & $0.00$ & $.0000$ & $.0000$ \\
\TaskNavNeg & $.27$ & $2.90$ & $1.24$ & $-.0041$ & $.0210$ \\
\TaskControl & $.29$ & $4.40$ & $2.04$ & $.0219$ & $.0028$ \\
\TaskLLM & $.30$ & $4.30$ & $2.07$ & $.0352$ & $.0149$ \\
\TaskNav & $.28$ & $4.50$ & $2.15$ & $.0661$ & $.0184$ \\
\TaskManip & $.35$ & $4.80$ & $2.47$ & $.0843$ & $.0219$ \\
\TaskStack & $.30$ & $5.00$ & $2.77$ & $.0951$ & $.0197$ \\
\TaskOGCube & $.31$ & $6.60$ & $4.19$ & $.2118$ & $-.0027$ \\
\TaskOGScene & $.29$ & $6.50$ & $4.05$ & $.1912$ & $.0033$ \\
\TaskRare & $.26$ & $2.77$ & $0.86$ & $-.0031$ & $.0065$ \\
\midrule
\TaskMGFour & $.27$ & $3.60$ & $1.43$ & $.1118$ & $.0071$ \\
\TaskOGPuzzle & $.33$ & $6.80$ & $4.60$ & $.2141$ & $-.0036$ \\
\midrule
\multicolumn{4}{l}{\emph{mean over 9 positive comparisons}} & $.0776$ & $.0118$ \\
\bottomrule
\end{tabular}
\end{table}

\paragraph{Depth reassignment.}
Depth reassignment would complement the equal-work control: permuting the recorded depth multiset over a common set of saved states preserves total refinement while removing the association between state and depth.
Its effect can be measured on the saved states under a common continuation policy, which isolates one decision at fixed occupancy, or with full rollouts, which include changes in future state visitation; in both cases the comparison is paired by seed and evaluation states.

\section{Critical-State Annotations and a PickCube Case Study}
\label{app:pr-full}
\label{app:semantic}
Semantic annotations test whether greater depth coincides with recognizable task events.
The proxies are torque-based recovery in HalfCheetah, flagged when joint torque exceeds the 90th percentile of the evaluation episodes; contact, gripper closure, lifting and stacking transitions in ManiSkill; junction proximity in AntMaze, within 1.5\,m of a corridor junction; slip recovery in RareEvent, flagged when cube--gripper relative speed exceeds 2\,cm/s; and goal-distance thresholds in OGBench, with bins at 0.05, 0.1 and 0.2\,m.
In ManiSkill, contact begins when either finger reports a contact force above 0.5\,N, grasp requires both fingers in contact with gripper closure above 80\%, lift begins when the cube rises 2\,cm above the table and stabilization/recovery covers the remaining lifted steps, including regrasps after a slip.
The PickCube phase study uses the five benchmark AIC checkpoints at $B=8$ with 50 evaluation episodes per seed, so its population matches the PickCube depth in \Cref{tab:halt-step-app}.

\begin{table}[h]
\centering\scriptsize
\caption{PickCube phase analysis over 25{,}000 evaluation decisions. Mean depth is mean $\pm$ standard deviation across five seeds; high-depth states satisfy $\tau\ge5$, the 75th-percentile depth, which selects 35.4\% of all decisions because of ties at the cutoff. Deeper-refinement gain is the mean change in normalized return-to-go when a state's executed depth is increased by two refinements; early-exit change is the corresponding change when the state is forced to depth one.}
\label{tab:critical-pr}
\setlength{\tabcolsep}{4pt}
\begin{tabular}{lrrrrr}
\toprule
Phase & Share & Mean depth & High-depth states & Deeper-refinement gain & Early-exit change\\
\midrule
Approach & 56.1\% & $1.8\pm.2$ & 12\% & $+.005$ & $-.004$\\
Grasp/contact & 14.6\% & $5.7\pm.4$ & 68\% & $+.046$ & $-.061$\\
Lift & 14.6\% & $4.9\pm.3$ & 54\% & $+.038$ & $-.047$\\
Stabilization/recovery & 14.7\% & $6.2\pm.5$ & 74\% & $+.053$ & $-.069$\\
\bottomrule
\end{tabular}
\end{table}
AIC spends the most refinement and gains the most from it, around contact and recovery, while the approach phase stays shallow.

\section{Task-Level Analysis and Cross-Task Association}
\label{app:regime-detail}
Across the six factorial tasks, stopping effects exceed supervision effects on PickCube, AntMaze, Memory, DoorKey and StackCube, where the adaptive configurations also execute more refinements, while RareEvent shows the reverse (\Cref{app:supervision-vs-allocation}).

Among the remaining tasks, FourRooms stays below its mean-depth-matched baseline throughout the sweep, Puzzle improves with stronger value supervision, DistShift is a saturated tie and HalfCheetah, low-heterogeneity AntMaze and RTE show small positive differences.
RareEvent pairs a shallow mean depth of $1.86$ with a large supervision effect.

\begin{figure}[t]
\centering
\includegraphics[width=\linewidth]{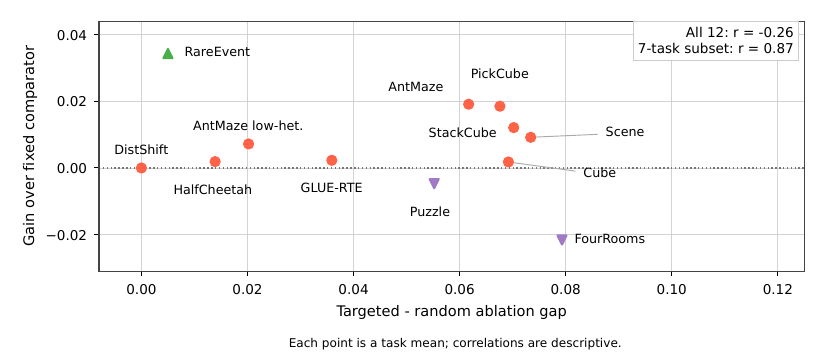}
\caption{\textbf{Allocation diagnostic and fixed-baseline gain.}
Each point is a task mean from the twelve-task benchmark. The seven-task subset (DistShift, HalfCheetah, low-heterogeneity AntMaze, RTE, AntMaze, PickCube, StackCube) has Pearson $r=0.87$; including Cube and Scene gives $r=0.62$ and all twelve points give $r=-0.26$. The horizontal axis is count-matched truncation sensitivity; the correlation depends on the included task set.}
\label{fig:teaser}
\end{figure}

\paragraph{Association across tasks.}
\Cref{fig:teaser} relates each task's targeted-minus-random drop to its gain over the per-budget fixed baseline.
The correlation is $r=.87$ on a seven-task subset, $.62$ after adding Cube and Scene and $-.26$ over all twelve tasks, so even its sign depends on the task set.
The count-matched difference also mixes state choice with removed work and the gain compares separately trained policies.
With the equal-work difference of \Cref{tab:equal-work} instead, the correlation is $r=.39$ over all twelve tasks and $r=.75$ on the seven-task subset.
Our conclusions rest on the per-task evidence.

\section{Limitations and Failure Modes}
\label{app:limitations}

\paragraph{Predictability and non-monotone refinement.}
State-dependent computation helps when useful additional depth can be predicted from the current representation.
Low immediate value gain can precede a larger later gain, invalidating a first-crossing heuristic without diminishing returns. The continuation recursion in \Cref{app:halt-derivation} handles this possibility; the shared-target critics do not verify its counterfactual value assumptions.

\paragraph{Shared targets and intervention interpretation.}
Shared GAE targets do not identify forced-depth counterfactual returns. The stated forward graph also makes a frozen policy invariant to value-output replacement (\Cref{prop:value-output-invariance}); such a replacement cannot test its stopping mechanism without changing additional parts of the policy.
Count-matched targeted/random truncation can remove different amounts of compute; the equal-work control in \Cref{tab:equal-work} matches removed refinements and reduces the mean targeted-minus-random difference from $.0463$ to $.0118$. A depth-reassignment control would further separate the association between state and depth.

\paragraph{Reward scale and operating point.}
The computation penalty depends on reward units. FourRooms remains below its matched baseline throughout the tested grid, whereas Puzzle improves under stronger value supervision. The default setting defines one operating point, with task-specific sensitivity to its loss weights.

\paragraph{Deployment versus training.}
Training takes $1.22\times$ the Fixed-8 time. End-to-end speedup is smaller when an expensive encoder dominates, as shown by GPT-2/RTE's 7.8\% latency reduction despite 49\% fewer refinement steps.
A depth cap bounds the number of refinement steps, but a hardware-specific latency guarantee also requires bounds on the encoder, heads, scheduling and execution time.

\paragraph{Broader impact.}
Adaptive computation can lower inference latency and energy for control policies when refinement is a substantial share of the per-decision cost; for policies dominated by a large frozen encoder the saving is small (\Cref{app:efficiency}) and training costs $1.22\times$ the Fixed-8 time.
Like other RL methods, AIC can be applied to systems in which errors cause physical harm and state-dependent stopping adds a specific failure mode: halting too early in rare states that the training data cover poorly.
Deployments in safety-critical settings should monitor stopping depth and can enforce a minimum depth where errors are costly.

\paragraph{Scale and reproducibility.}
The experiments use small refinement modules and simulated environments; scaling to large backbones and real robots remains untested.
The privileged reference is a diagnostic rather than a certified upper bound.
The factorial, timing and sensitivity studies use run sets separate from the twelve-task benchmark; the code release includes their seed-level records and compute accounting.

\section{Memory Task Extensions}
\label{app:pomdp}

Memory and DoorKey extend AIC to decisions that depend on past observations, which require a history representation in addition to within-decision refinement (\Cref{tab:pomdp}).

\begin{table}[h]
\centering
\caption{\textbf{Memory tasks at $B = 8$.}
Fixed uses depth eight on Memory and DoorKey; PickCube and AntMaze repeat the baselines in \Cref{tab:main}.
These runs are independent of the stopping-by-supervision factorial. P/R: precision and recall of high-depth decisions at critical states (\Cref{app:pomdp}).}
\label{tab:pomdp}
\scriptsize
\setlength{\tabcolsep}{4pt}
\begin{tabular}{@{}l cc cc c c c c@{}}
\toprule
& \multicolumn{2}{c}{AIC} & \multicolumn{2}{c}{ACT-ponder} & non-VoC & Fixed & Tgt$-$Rnd & P/R \\
\cmidrule(lr){2-3}\cmidrule(lr){4-5}
Task & Rew. & $\bar d$ & Rew. & $\bar d$ & Rew. & Rew. & gap & (crit.\ states) \\
\midrule
MiniGrid \textit{Memory}  & $.5715$ & $3.31$ & $.5690$ & $3.62$ & $.5677$ & $.5504$ & $+.056$ & $.42 / .79$ \\
MiniGrid \textit{DoorKey} & $.5470$ & $3.36$ & $.5413$ & $3.86$ & $.5431$ & $.5259$ & $+.059$ & $.44 / .85$ \\
\midrule
\multicolumn{9}{l}{\emph{Benchmark tasks at $B=8$, for scale.}} \\
\TaskManip   & $.4926$ & $3.47$ & $.4910$ & $3.80$ & $.4907$ & $.4741$ & $+.068$ & $.49 / .74$ \\
\TaskNav     & $.5209$ & $3.15$ & $.5200$ & $3.54$ & $.5208$ & $.5018$ & $+.062$ & $.41 / .77$ \\
\bottomrule
\end{tabular}
\end{table}

\paragraph{Tasks.}
We use two MiniGrid environments with partial observability~\citep{chevalier2018minimalistic}.
In \textit{Memory}, the agent enters a corridor where a colored cue is shown, then proceeds to a junction where it must select the path matching the cue; the cue is no longer visible at the junction, so the policy must internalize it across timesteps.
In \textit{DoorKey}, the agent must locate a key, pick it up, traverse to a locked door and use the key; the key and door are typically not co-visible in the agent's egocentric view.
Both tasks use sparse completion rewards and require the representation to retain relevant history.

\paragraph{History representation.}
Both tasks use a single-layer GRU with 256 hidden units over the CNN embedding of each observation and the previous action.
The GRU state $\psi_t$ is updated once per environment step and $z_0=f_{\mathrm{enc}}(\psi_t)$; refinement does not modify $\psi_t$ and $z_\tau$ is not fed back into the GRU.
Training uses truncated backpropagation through time over 32 environment steps and ACT-ponder, learned non-VoC and Fixed-8 share the same GRU and encoder.

\paragraph{Critical-state precision and recall.}
A decision is high-depth when $\tau$ is at or above its seed's 75th-percentile depth.
The critical states are the simulator-annotated junction step in Memory, where the cue must be recalled and the key-pickup and door-opening steps in DoorKey.
Precision is the fraction of high-depth decisions within one step of a critical state and recall is the fraction of critical states whose decision is high-depth.
For the benchmark rows of \Cref{tab:pomdp}, a decision is critical when it lies within one step of contact onset, grasp closure, lift-off or slip recovery in PickCube, which covers 23\% of decisions, or within 1.5\,m of a corridor junction in AntMaze, which covers 15\%; precision and recall are computed over these critical decisions.
These event windows are narrower than the phases of \Cref{tab:critical-pr}: treating whole grasp, lift and stabilization phases as critical would give PickCube a precision of .81 and a recall of .65 instead of .49 and .74.

\paragraph{Results.}
The factorial has larger stopping-configuration effects on both tasks (\Cref{app:regime-detail}), where the adaptive configurations execute 3.3--3.5 refinements against one for the non-adaptive configurations.
On Memory, AIC obtains reward $0.5715$ at mean depth $\bar d=3.31$, a $+0.021$ increase over Fixed-$8$ with $0.31$ fewer steps than ACT-ponder.
On DoorKey, it obtains $0.5470$ at $\bar d=3.36$, again with lower mean depth.
The targeted-minus-random gaps are $+0.056$ and $+0.059$, respectively.
Against simulator annotations, precision and recall are $(0.42,0.79)$ at memory-query positions and $(0.44,0.85)$ at key-pickup positions: most critical states receive high depth, while fewer than half of the high-depth decisions fall at critical states.

Memory and DoorKey are not part of the twelve-task benchmark or the cross-task correlation.

\section{Training Cost and Measured Inference Latency}
\label{app:efficiency}
The timing study separates the cost of training all intermediate heads from the savings of stopping early at inference. It also distinguishes refinement count from end-to-end runtime.
All timings use a single NVIDIA A100 80\,GB GPU with an AMD EPYC 7763 host CPU, PyTorch 2.3 and CUDA 12.1.
Inference latency is the median over 1{,}000 decisions after 100 warm-up decisions at batch size one; the control and robotics rows use the five benchmark checkpoints at $B=8$ and the RTE row uses the timing checkpoint described in \Cref{app:tasks-llm}.
The 31--44\% reductions are therefore per-decision latency, not throughput.

\begin{table}[h]
\centering
\caption{\textbf{Timing and computation.} Inference latency is the median per-decision time at batch size one; training time is wall-clock time for the full run. Mean steps are refinements per decision.}
\label{tab:efficiency}
\scriptsize
\setlength{\tabcolsep}{5pt}
\begin{tabular}{@{}l cc c c@{}}
\toprule
Setting & Fixed-8 & AIC & Mean steps (Fixed-8 / AIC) & Change \\
\midrule
GPT-2/RTE inference & 9.41\,ms & 8.68\,ms & 8 / 4.08 & $-7.8\%$ latency \\
HalfCheetah inference & 0.86\,ms & 0.48\,ms & 8 / 3.04 & $-44\%$ latency \\
AntMaze inference & 0.89\,ms & 0.53\,ms & 8 / 3.15 & $-40\%$ latency \\
PickCube inference & 0.92\,ms & 0.57\,ms & 8 / 3.47 & $-38\%$ latency \\
StackCube inference & 0.94\,ms & 0.65\,ms & 8 / 3.77 & $-31\%$ latency \\
PickCube training & 5.9\,h & 7.2\,h & all eight depths & $1.22\times$ time \\
\bottomrule
\end{tabular}
\end{table}

The training comparison holds hardware, batch configuration, update count and training setup fixed; both PickCube runs use ten million environment steps.
Every refinement depth is evaluated during optimization for value supervision, including depths beyond the sampled action.
Fixed-8 evaluates the same eight refinement steps and value heads, so the additional AIC time comes from the halting heads, the survival and joint log-probability computations and the per-sample gather of the selected depth's action head, which our implementation does not fuse into one kernel.
At inference, total work includes the encoder, executed refinements, heads and control-flow overhead.
Refinement count and total FLOPs therefore decrease by the same proportion only when the remaining costs are negligible.
For an additive per-action cost model $T(d)=T_{\rm enc}+dT_{\rm ref}+T_{\rm other}$ with constant marginal refinement time, the fractional time saved by reducing mean depth from $d_0$ to $d_1$ is
\begin{equation}
\frac{T(d_0)-T(d_1)}{T(d_0)}
=\frac{(d_0-d_1)T_{\rm ref}}{T_{\rm enc}+d_0T_{\rm ref}+T_{\rm other}}.
\end{equation}
Fitting this model to \Cref{tab:efficiency} gives $T_{\rm ref}\approx0.19$\,ms and $T_{\rm enc}+T_{\rm other}\approx7.92$\,ms for RTE, so refinement accounts for about 16\% of Fixed-8's per-decision time and $T_{\rm ref}\approx0.07$--$0.08$\,ms with $0.25$--$0.39$\,ms of fixed overhead for the control and robotics policies.
The encoder and other fixed costs explain how a large refinement reduction can produce a smaller end-to-end change.
The experiments in this paper used about 9{,}800 A100 GPU-hours in total, plus about 1{,}200 GPU-hours of preliminary and discarded runs.

\section{Base-Policy Transfer Details}
\label{app:transfer}
\Cref{tab:base_policy_task_gain} summarizes the eight-task transfer to DiffusionQL, DPPO, D3P, DSAC-D and IDQL~\citep{wang2022diffusion,dppo,d3p,dsac,idql}.
Each policy family defines its own refinement operation, intermediate-action decoder and training update, so one step can have a different computational cost across families.
\Cref{tab:transfer-spec} specifies how AIC is attached to each family and \Cref{tab:transfer-full} lists the full results.
In every family one refinement is one denoising step of an eight-step sampler, the halting head reads the denoiser's hidden features at that step concatenated with the state embedding and the intermediate action is the one-step prediction $\hat a_0$ of the clean action from the partially denoised sample.
DPPO and D3P are trained on-policy, so their halting heads use the joint PPO ratio of \Cref{sec:method} with the denoising chain as internal refinement.
DiffusionQL and IDQL are trained offline and DSAC-D off-policy, where no rollout ratio is available; their halting heads instead maximize $\E_{\tau\sim q_\theta}[Q(s,\hat a_{0,\tau})]-\lambda_c\E[\tau]/\dmax$ with a score-function gradient through $q_\theta$, holding the family's own critic fixed during the halting update.
D3P's own adaptive step allocation is disabled in both its Fixed and AIC variants, so the comparison isolates the AIC halting head.

\paragraph{Datasets and discrete actions.}
DiffusionQL and IDQL train on fixed datasets of one million transitions per task: the D4RL \texttt{halfcheetah-medium-expert-v2} dataset, uniformly subsampled from its two million transitions, the full D4RL \texttt{antmaze-medium-diverse-v2} dataset, the OGBench \texttt{scene-play-v0} and \texttt{puzzle-4x4-play-v0} datasets and, for ManiSkill, MiniGrid and RareEvent, which have no standard offline datasets, transitions collected by the benchmark Fixed-8 policy with Gaussian action noise of standard deviation $0.2$, or $20\%$ uniformly random actions on FourRooms.
The D4RL \texttt{-v2} data were recorded in the mujoco-py version of HalfCheetah; HalfCheetah-v4 keeps its observation and action spaces, so we evaluate the trained policies in HalfCheetah-v4 and normalize returns with D4RL's reference scores.
DPPO and D3P fine-tune online from a behavior-cloned initialization on the same data and DSAC-D trains online from its replay buffer.
For FourRooms, whose seven actions are discrete, every family denoises a seven-dimensional continuous vector and executes its argmax.

\begin{table}[h]
\centering
\caption{\textbf{Attaching AIC to each policy family.} One refinement is one denoising step; the intermediate action is the one-step clean-action prediction $\hat a_0$.}
\label{tab:transfer-spec}
\scriptsize
\setlength{\tabcolsep}{4pt}
\begin{tabular}{@{}l l l l c@{}}
\toprule
Family & Training regime & Halting update & Critic or baseline & Fixed steps \\
\midrule
DiffusionQL & offline & critic-weighted, score function & DiffusionQL double-$Q$ critic & 8 \\
IDQL & offline & critic-weighted, score function & IQL $Q$-function & 8 \\
DSAC-D & off-policy & critic-weighted, score function & mean of the distributional critic & 8 \\
DPPO & on-policy & joint PPO ratio & GAE with state-only baseline & 8 \\
D3P & on-policy & joint PPO ratio & GAE with state-only baseline & 8 \\
\bottomrule
\end{tabular}
\end{table}

\begin{table}[h]
\centering
\caption{\textbf{Full transfer results at nominal $B=8$.}
Fixed and AIC give mean return with the standard deviation over five seeds as a subscript; Gain is $\Delta = R_{\mathrm{AIC}} - R_{\mathrm{Fixed}}$ with its standard error and AIC Steps is the mean halting depth with its standard deviation.
Fixed executes eight refinement steps per action in every family.}
\label{tab:transfer-full}
\fontsize{8}{8.5}\selectfont
\setlength{\tabcolsep}{3.5pt}
\begin{tabular}{llcccc}
\toprule
Base Policy & Task / Env & Fixed & AIC & Gain & AIC Steps \\
\midrule
DiffusionQL & HalfCheetah-v4 & $.934_{\pm.004}$ & $.936_{\pm.004}$ & $+.002_{\pm.003}$ & $2.8_{\pm.2}$ \\
DiffusionQL & ManiSkill PickCube-v1 & $.505_{\pm.005}$ & $.525_{\pm.005}$ & $+.020_{\pm.003}$ & $3.8_{\pm.3}$ \\
DiffusionQL & ManiSkill StackCube-v1 & $.486_{\pm.004}$ & $.501_{\pm.006}$ & $+.015_{\pm.003}$ & $4.1_{\pm.3}$ \\
DiffusionQL & OGBench scene-play-v0 & $.430_{\pm.005}$ & $.442_{\pm.004}$ & $+.012_{\pm.003}$ & $6.2_{\pm.2}$ \\
DiffusionQL & AntMaze main & $.520_{\pm.004}$ & $.537_{\pm.004}$ & $+.017_{\pm.003}$ & $3.6_{\pm.2}$ \\
DiffusionQL & OGBench puzzle-4x4-play-v0 & $.421_{\pm.004}$ & $.418_{\pm.005}$ & $-.003_{\pm.003}$ & $7.2_{\pm.4}$ \\
DiffusionQL & MiniGrid-FourRooms & $.970_{\pm.008}$ & $.955_{\pm.009}$ & $-.015_{\pm.005}$ & $2.6_{\pm.1}$ \\
DiffusionQL & RareEvent-StackCube-Misalign & $.590_{\pm.008}$ & $.615_{\pm.009}$ & $+.025_{\pm.005}$ & $1.8_{\pm.1}$ \\
\midrule
DPPO & HalfCheetah-v4 & $.933_{\pm.003}$ & $.934_{\pm.003}$ & $+.001_{\pm.002}$ & $2.6_{\pm.2}$ \\
DPPO & ManiSkill PickCube-v1 & $.492_{\pm.006}$ & $.510_{\pm.005}$ & $+.018_{\pm.003}$ & $3.5_{\pm.2}$ \\
DPPO & ManiSkill StackCube-v1 & $.471_{\pm.005}$ & $.484_{\pm.005}$ & $+.013_{\pm.003}$ & $3.8_{\pm.1}$ \\
DPPO & OGBench scene-play-v0 & $.417_{\pm.004}$ & $.425_{\pm.005}$ & $+.008_{\pm.003}$ & $5.6_{\pm.2}$ \\
DPPO & AntMaze main & $.510_{\pm.004}$ & $.525_{\pm.003}$ & $+.015_{\pm.002}$ & $3.3_{\pm.3}$ \\
DPPO & OGBench puzzle-4x4-play-v0 & $.410_{\pm.004}$ & $.405_{\pm.004}$ & $-.005_{\pm.003}$ & $6.8_{\pm.4}$ \\
DPPO & MiniGrid-FourRooms & $.968_{\pm.008}$ & $.950_{\pm.008}$ & $-.018_{\pm.005}$ & $2.5_{\pm.2}$ \\
DPPO & RareEvent-StackCube-Misalign & $.575_{\pm.009}$ & $.597_{\pm.007}$ & $+.022_{\pm.005}$ & $1.9_{\pm.2}$ \\
\midrule
D3P & HalfCheetah-v4 & $.935_{\pm.004}$ & $.937_{\pm.004}$ & $+.002_{\pm.003}$ & $2.9_{\pm.2}$ \\
D3P & ManiSkill PickCube-v1 & $.515_{\pm.005}$ & $.538_{\pm.004}$ & $+.023_{\pm.003}$ & $4.0_{\pm.2}$ \\
D3P & ManiSkill StackCube-v1 & $.496_{\pm.005}$ & $.514_{\pm.005}$ & $+.018_{\pm.003}$ & $4.3_{\pm.3}$ \\
D3P & OGBench scene-play-v0 & $.438_{\pm.004}$ & $.452_{\pm.005}$ & $+.014_{\pm.003}$ & $6.5_{\pm.3}$ \\
D3P & AntMaze main & $.525_{\pm.004}$ & $.543_{\pm.004}$ & $+.018_{\pm.003}$ & $3.8_{\pm.3}$ \\
D3P & OGBench puzzle-4x4-play-v0 & $.420_{\pm.004}$ & $.422_{\pm.004}$ & $+.002_{\pm.003}$ & $7.4_{\pm.2}$ \\
D3P & MiniGrid-FourRooms & $.969_{\pm.007}$ & $.957_{\pm.007}$ & $-.012_{\pm.004}$ & $2.8_{\pm.3}$ \\
D3P & RareEvent-StackCube-Misalign & $.600_{\pm.008}$ & $.630_{\pm.009}$ & $+.030_{\pm.005}$ & $2.0_{\pm.1}$ \\
\midrule
DSAC-D & HalfCheetah-v4 & $.934_{\pm.004}$ & $.935_{\pm.004}$ & $+.001_{\pm.003}$ & $2.2_{\pm.1}$ \\
DSAC-D & ManiSkill PickCube-v1 & $.485_{\pm.006}$ & $.495_{\pm.006}$ & $+.010_{\pm.004}$ & $3.1_{\pm.3}$ \\
DSAC-D & ManiSkill StackCube-v1 & $.462_{\pm.005}$ & $.470_{\pm.005}$ & $+.008_{\pm.003}$ & $3.4_{\pm.1}$ \\
DSAC-D & OGBench scene-play-v0 & $.405_{\pm.005}$ & $.409_{\pm.005}$ & $+.004_{\pm.003}$ & $5.0_{\pm.2}$ \\
DSAC-D & AntMaze main & $.500_{\pm.004}$ & $.510_{\pm.004}$ & $+.010_{\pm.003}$ & $3.0_{\pm.2}$ \\
DSAC-D & OGBench puzzle-4x4-play-v0 & $.407_{\pm.003}$ & $.401_{\pm.003}$ & $-.006_{\pm.002}$ & $6.5_{\pm.2}$ \\
DSAC-D & MiniGrid-FourRooms & $.970_{\pm.006}$ & $.950_{\pm.008}$ & $-.020_{\pm.004}$ & $2.3_{\pm.2}$ \\
DSAC-D & RareEvent-StackCube-Misalign & $.560_{\pm.009}$ & $.572_{\pm.008}$ & $+.012_{\pm.005}$ & $1.6_{\pm.2}$ \\
\midrule
IDQL & HalfCheetah-v4 & $.933_{\pm.003}$ & $.934_{\pm.003}$ & $+.001_{\pm.002}$ & $2.4_{\pm.1}$ \\
IDQL & ManiSkill PickCube-v1 & $.490_{\pm.004}$ & $.502_{\pm.006}$ & $+.012_{\pm.003}$ & $3.3_{\pm.2}$ \\
IDQL & ManiSkill StackCube-v1 & $.468_{\pm.005}$ & $.478_{\pm.005}$ & $+.010_{\pm.003}$ & $3.6_{\pm.3}$ \\
IDQL & OGBench scene-play-v0 & $.424_{\pm.004}$ & $.434_{\pm.005}$ & $+.010_{\pm.003}$ & $6.0_{\pm.3}$ \\
IDQL & AntMaze main & $.506_{\pm.004}$ & $.520_{\pm.004}$ & $+.014_{\pm.003}$ & $3.4_{\pm.3}$ \\
IDQL & OGBench puzzle-4x4-play-v0 & $.413_{\pm.004}$ & $.414_{\pm.004}$ & $+.001_{\pm.003}$ & $7.0_{\pm.4}$ \\
IDQL & MiniGrid-FourRooms & $.969_{\pm.006}$ & $.955_{\pm.008}$ & $-.014_{\pm.004}$ & $2.5_{\pm.2}$ \\
IDQL & RareEvent-StackCube-Misalign & $.570_{\pm.008}$ & $.588_{\pm.008}$ & $+.018_{\pm.005}$ & $1.7_{\pm.2}$ \\
\bottomrule
\end{tabular}
\end{table}

\section{Experimental Protocol Summary}
\label{app:inventory}
The main benchmark covers twelve tasks at four nominal budgets with five seeds each and the POGP comparison covers four environments with five seeds under an eight-step cap.
The transfer study covers eight tasks in each of five policy families; the memory-task comparisons and the stopping-by-supervision factorial, four configurations on six tasks, use their own runs.
The sensitivity study covers five-by-four loss-weight grids for FourRooms and Puzzle.
The truncation interventions reuse the frozen benchmark checkpoints and average ten random subsets per seed for both the count-matched and equal-work controls.
The local-mask comparison uses five fresh seeds and the entropy/probe and timing studies use their own runs.

\paragraph{Uncertainty.}
Each caption states its uncertainty convention.
The benchmark, the halting-baseline, transfer and POGP comparisons and the data-source and stopping and supervision ablations give standard deviations over five seeds and the sensitivity grids give standard errors over five seeds.
RTE accuracies average ten sampled evaluation passes per seed before the across-seed statistics (\Cref{app:tasks-llm}).
Correlations are computed from task means.

\paragraph{Protocol details.}
The fixed-baseline selection rule and seed counts (\Cref{sec:matched,app:matched-full}), the random-stopping law (\Cref{app:baseline-defs}), the halting-baseline specifications (\Cref{tab:halt-spec}), factorial schedules and depths (\Cref{tab:alloc-vs-sup}), annotation thresholds and populations (\Cref{app:semantic}), memory and transfer interfaces (\Cref{app:pomdp,app:transfer}), budget-specific reference rewards (\Cref{tab:full-budgets}) and timing hardware and durations (\Cref{app:efficiency}) accompany the corresponding results.
\Cref{app:limitations} discusses the limitations.

\clearpage

\end{document}